\documentclass{article}

\PassOptionsToPackage{numbers, compress}{natbib}

\usepackage[final]{neurips_2023}

\usepackage{nicefrac}       %

\usepackage{scalefnt,letltxmacro}
\LetLtxMacro{\oldtextsc}{\textsc}
\renewcommand{\textsc}[1]{\oldtextsc{\scalefont{1.10}#1}}
\usepackage[acronym,nowarn]{glossaries}
\setacronymstyle{long-sc-short}
\glsdisablehyper
\makeglossaries
\usepackage{xspace}
\usepackage{float}
\usepackage{physics}

\usepackage{enumitem}

\usepackage{wrapfig}

\usepackage{amssymb}
\usepackage{mathtools}
\usepackage{amsfonts}
\usepackage{amsmath}
\usepackage{amsthm}
\usepackage{booktabs}
\usepackage[]{microtype}

\usepackage[linesnumbered,ruled,vlined]{algorithm2e}

\SetCommentSty{mycommfont}
\SetKwInput{KwInput}{Input} 
\SetKwInput{KwOutput}{Output}

\usepackage[colorlinks,linktoc=all]{hyperref}
\usepackage[all]{hypcap}
\hypersetup{citecolor=Mahogany}
\hypersetup{linkcolor=Mahogany}
\hypersetup{urlcolor =Mahogany}

\usepackage[capitalize,nameinlink]{cleveref}
\crefname{section}{\S}{\S\S}
\Crefname{section}{\S}{\S\S}
\creflabelformat{equation}{#2\textup{#1}#3}

\usepackage[usenames,dvipsnames,svgnames,table]{xcolor}
\definecolor{shadecolor}{gray}{0.9}

\definecolor{mylightgray}{gray}{0.94}

\definecolor{my_blue}{HTML}{468284}
\definecolor{my_red}{HTML}{8b0000}

\definecolor{negative_red}{HTML}{990000}
\definecolor{positive_green}{HTML}{009900}

\usepackage{algcompatible}

\makeatletter
\newcommand\fs@coloruled{\def\@fs@cfont{\bfseries}\let\@fs@capt\floatc@ruled
  \def\@fs@pre{{\color{blue}\hrule height.8pt depth0pt }\kern2pt}%
  \def\@fs@post{\kern2pt{\color{blue}\hrule}\relax}%
  \def\@fs@mid{\kern2pt{\color{blue}\hrule}\kern2pt}%
  \let\@fs@iftopcapt\iftrue}
\makeatother
\makeatletter
\@ifundefined{c@rownum}{%
  \let\c@rownum\rownum
}{}
\@ifundefined{therownum}{%
  \def\therownum{\@arabic\rownum}%
}{}
\makeatother

\makeatletter
\newcommand*{\addFileDependency}[1]{%
	\typeout{(#1)}
	\@addtofilelist{#1}
	\IfFileExists{#1}{}{\typeout{No file #1.}}
}
\makeatother

\usepackage[font=small,labelfont=bf,tableposition=top]{caption}
\usepackage{tikz}
\usepackage{graphicx}
\usepackage[]{subcaption}   %

\usepackage{pgfplots}
\pgfplotsset{compat=1.6}
\usepgfplotslibrary{groupplots}
\tikzstyle{every picture}+=[font=\sffamily]
\tikzstyle{optimized} = [circle,fill=white,draw=black, dashed,inner sep=1pt, minimum size=20pt, font=\fontsize{10}{10}\selectfont, node distance=1]
\pgfkeys{/pgf/number format/.cd,1000 sep={}}
\pgfplotsset{
	tick label style = {font=\sffamily},
	every axis label/.append style={font=\sffamily},
	typeset ticklabels with strut,
}
\pgfplotsset{every axis/.append style={
			every x tick label/.append style={font=\fontsize{6pt}{6pt}\sffamily, yshift=.5ex,},
			every y tick label/.append style={font=\fontsize{6pt}{6pt}\sffamily, xshift=.5ex},
			every y label/.append style={xshift=10ex, font=\sffamily},
			every x label/.append style={yshift=3ex, font=\sffamily},
			every title/.append style={font=\sffamily}
		},
}
\pgfplotsset{
  xticklabel={$\mathsf{\pgfmathprintnumber{\tick}}$},
  yticklabel={$\mathsf{\pgfmathprintnumber{\tick}}$},
}
\pgfplotsset{every axis title/.append style={yshift=-1ex}}
\newlength\figureheight
\newlength\figurewidth
\usepgfplotslibrary{external}
\usepackage[colorinlistoftodos,
	textsize=scriptsize,
	linecolor=red!30,
	bordercolor=red!30,
	backgroundcolor=red!10]{todonotes}
\renewcommand{\todo}[2][]{\tikzexternaldisable\@todo[#1]{#2}\tikzexternalenable}

\usepackage[most]{tcolorbox}
{\begin{tcolorbox}[toprule=2mm,left=4pt,right=4pt,top=0pt,bottom=1pt,boxsep=2pt, before skip = 2ex, after skip =2ex]}{\end{tcolorbox}}
\tcbset{boxsep=4pt,left=2pt,right=2pt,top=-0pt,bottom=0pt}

\usepackage{url}

\usepackage{xr}
\usepackage{subcaption}

\usepackage{tikz}
\usepackage{graphicx}
\usepackage{xcolor}
\usepackage{amsmath}
\usepackage{amssymb}
\usepackage{bm}
\usepackage{amsthm} %
\newtheorem{theorem}{Theorem}

\newtheorem{definition}{Definition}

\usepackage{adjustbox}
\usepackage{multirow}
\usepackage{mathtools}

\usepackage{xspace}

\newacronym{MAP}{map}{maximum-a-posteriori}
\newacronym{MLE}{mle}{maximum likelihood estimation}
\newacronym{MNLL}{mnll}{mean negative loglikelihood}
\newacronym{NLL}{nll}{negative loglikelihood}
\newacronym{LL}{ll}{log-likelihood}
\newacronym{RMSE}{rmse}{root mean square error}
\newacronym{ECE}{ece}{expected calibration error}
\newacronym{FID}{fid}{Fr\'echet Inception Distance}

\newacronym{AE}{ae}{autoencoder}
\newacronym{WAE}{wae}{Wasserstein Autoencoder}
\newacronym{VAE}{vae}{Variational Autoencoder}
\newacronym{BAE}{bae}{Bayesian autoencoder}
\newacronym{CDF}{cdf}{cumulative density function}
\newacronym{GAN}{gan}{Generative Adversarial Network}
\newacronym{GM}{gm}{Generative Model}

\newacronym{PDE}{pde}{partial differential equation}
\newacronym{DCT}{dct}{discrete cosine transformation}

\newacronym{MC}{mc}{Monte Carlo}
\newacronym{MCMC}{mcmc}{Markov chain Monte Carlo}
\newacronym{HMC}{hmc}{Hamiltonian Monte Carlo}
\newacronym{MH}{mh}{Metropolis-Hastings}
\newacronym{NUTS}{nuts}{no-u-turn sampler}
\newacronym{SGHMC}{sghmc}{stochastic gradient Hamiltonian Monte Carlo}

\newacronym{MMD}{mmd}{Maximum Mean Discrepancy}
\newacronym{GMM}{gmm}{Gaussian Mixture Model}
\newacronym{EBM}{ebm}{Energy-Based Model}
\newacronym{ARM}{arm}{Autoregressive Model}
\newacronym{AVB}{avb}{Adversarial Varitional Bayes}

\newacronym{DGP}{dgp}{deep Gaussian process} %
\newacronym{GPLVM}{gplvm}{Gaussian process latent variable model}
\newacronym{DPMM}{dpmm}{Dirichlet Process Mixture Model}

\newacronym{VFE}{vfe}{variational free energy}

\newacronym[firstplural=Gaussian Processes]{GP}{gp}{Gaussian Process}

\newacronym{VI}{vi}{variational inference}

\newacronym{ELBO}{elbo}{evidence lower bound}
\newacronym{NELBO}{nelbo}{negative evidence lower bound}
\newacronym{ELL}{ell}{expected log likelihood}
\newacronym{KL}{kl}{Kullback-Leibler divergence}
\newacronym{AUC}{auc}{area under the curve}

\newacronym[firstplural=Bayesian neural networks]{BNN}{bnn}{Bayesian neural network}
\newacronym[firstplural=deep neural networks]{DNN}{dnn}{deep neural network}
\newacronym[]{CNN}{cnn}{convolutional neural network}
\newacronym{MLP}{mlp}{multilayer perceptron}
\newacronym{NN}{nn}{neural network}
\newacronym{RELU}{ReLU}{rectified linear unit}

\newacronym{NF}{nf}{Normalizing Flow}
\newacronym{DM}{dm}{Diffusion Model}

\newacronym{RBF}{rbf}{radial basis function}
\newacronym{ARD}{ard}{automatic relevance determination}

\newacronym{RKHS}{rkhs}{reproducing kernel Hilbert space}
\newacronym{SNR}{snr}{signal-to-noise ratio}
\newacronym{PDF}{pdf}{probability density function}

\newacronym{OT}{ot}{optimal transport}
\newacronym{WD}{wd}{Wasserstein distance}
\newacronym{SWD}{swd}{sliced-Wasserstein distance}
\newacronym{DSWD}{dswd}{distributional sliced-Wasserstein distance}

\newcommand{\iid}{i.i.d~} 

\newcommand{\name}[1]{{\textsc{#1}}\xspace}

\newcommand{\cifar}{\textsc{cifar}\textsc{\footnotesize{10}}\xspace}

\newcommand{\celeba}{\name{celeba}}

\newcommand{\mathbold}[1]{{\boldsymbol{\mathbf{#1}}}}

\newcommand{\g}{\,|\,}

\newcommand{\nestedmathbold}[1]{{\mathbold{#1}}}

\newcommand{\mbf}{\nestedmathbold{f}}

\newcommand{\mbh}{\nestedmathbold{h}}

\newcommand{\mbu}{\nestedmathbold{u}}

\newcommand{\mbx}{\nestedmathbold{x}}
\newcommand{\mby}{\nestedmathbold{y}}
\newcommand{\mbz}{\nestedmathbold{z}}

\newcommand{\mbA}{\nestedmathbold{A}}

\newcommand{\mbD}{\nestedmathbold{D}}

\newcommand{\mbI}{\nestedmathbold{I}}

\newcommand{\mbV}{\nestedmathbold{V}}

\newcommand{\mbphi}{\nestedmathbold{\phi}}

\newcommand{\mbpsi}{\nestedmathbold{\psi}}

\newcommand{\mbtheta}{\nestedmathbold{\theta}}

\newcommand{\mbLambda}{\nestedmathbold{\Lambda}}

\newcommand{\mbSigma}{\nestedmathbold{\Sigma}}

\DeclareRobustCommand{\KL}[2]{\ensuremath{\textsc{kl}\left[#1\;\|\;#2\right]}}
\DeclarePairedDelimiterX{\infdivx}[2]{[}{]}{%
  #1\;\delimsize\|\;#2%
}

\newcommand{\cD}{\mathcal{D}}
\newcommand{\cL}{\mathcal{L}}
\newcommand{\cN}{\mathcal{N}}
\newcommand{\cM}{\mathcal{M}}

\newcommand{\cH}{\mathcal{H}}

\title{One-Line-of-Code Data Mollification Improves Optimization of Likelihood-based Generative Models}

\author{%
  Ba-Hien Tran \\
  Department of Data Science\\
  EURECOM, France \\
  \texttt{ba-hien.tran@eurecom.fr} \\
  \And
  Giulio Franzese \\
  Department of Data Science \\
  EURECOM, France \\
  \texttt{giulio.franzese@eurecom.fr} \\
  \AND
  Pietro Michiardi \\
  Department of Data Science \\
  EURECOM, France \\
  \texttt{pietro.michiardi@eurecom.fr} \\
  \And
  Maurizio Filippone \\
  Department of Data Science \\
  EURECOM, France \\
  \texttt{maurizio.filippone@eurecom.fr} \\
}

\begin{document}

\maketitle

\begin{abstract}
	\glspl{GM} have attracted considerable attention due to their tremendous success in various domains, such as computer vision where they are capable to generate impressive realistic-looking images.
Likelihood-based \glspl{GM} are attractive due to the possibility to generate new data by a single model evaluation.
However, they typically achieve lower sample quality compared to state-of-the-art score-based \glspl{DM}.
This paper provides a significant step in the direction of addressing this limitation.
The idea is to borrow one of the strengths of score-based \glspl{DM}, which is the ability to perform accurate density estimation in low-density regions and to address manifold overfitting by means of data mollification.
{ We propose a view of data mollification within likelihood-based \glspl{GM} as a continuation method, whereby the optimization objective smoothly transitions from simple-to-optimize to the original target.}
Crucially, data mollification can be implemented by adding one line of code in the optimization loop, and we demonstrate that this provides a boost in generation quality of likelihood-based \glspl{GM}, without computational overheads.
We report results on real-world image data sets and \textsc{uci} benchmarks  with popular likelihood-based \glspl{GM}, including variants of variational autoencoders and normalizing flows, showing large improvements in \textsc{fid} score and density estimation.

\end{abstract}

\section{Introduction}

\glsreset{GM}\glspl{GM} have attracted considerable attention recently due to their tremendous success in various domains, such as computer vision, graph generation, physics and reinforcement learning \cite[see e.g.,][and references therein]{KingmaW19,Kobyzev2021,Tran21,Tran23}. %
Given a set of data points, \glspl{GM} attempt to characterize the distribution of such data so that it is then possible to draw new samples from this.
Popular approaches include \glspl{VAE}, \glspl{NF}, \glspl{GAN}, and score-based \glsreset{DM}\glspl{DM}.

{
  In general, the goal of any \glspl{GM} is similar to that of density estimation with the additional aim to do so by constructing a parametric mapping between an easy-to-sample-from distribution $p_{\mathrm{s}}$ and the desired data distribution $p_{\text{data}}$.
  While different \glspl{GM} approaches greatly differ in their optimization strategy and formulation, the underlying objectives share some similarity due to their relation to the optimal transport problem, defined as $\arg\min_{\pi}\int \norm{\mbx-\mby}^2d\pi(\mbx,\mby)$.
  Here $\pi$ is constrained to belong to the set of joint distributions with marginals $p_{\mathrm{s}},p_{\text{data}}$, respectively~\citep{Genevay2017arXiv,LiutkusICML2019}. %
  This unified perspective is explicitly investigated for \glspl{GAN} and \glspl{VAE}~\citep{Genevay2017arXiv} for example, whereas other works study \glspl{NF}~\citep{onken2021ot}.
  Similarly, \glspl{DM} can be connected to Schrodinger Bridges~\citep{chen2021likelihood}, which solve the problem of \textit{entropy-regularized} optimal transport~\citep{pavon2021data}.
  Given that extensions of the regularized optimal transport case are available also for other generative models~\cite{sanjabi2018convergence,reshetova2021understanding}, we should expect that, in principle, any technique should allow generation of samples with similar quality, provided it is properly tuned.
  However, this is not true in practice.
  The different formulations lead to a variety of properties associated with \glspl{GM}, and pros and cons of each formulation can be understood through the so-called \gls{GM} tri-lemma~\citep{XiaoICLR2022}.
  The three desirable properties of \glspl{GM} are high sample quality, mode coverage, and fast sampling, and it has been argued that such goals are difficult to be satisfied simultaneously \citep{XiaoICLR2022} .
}

The state-of-the-art is currently dominated by score-based \glspl{DM}, due to their ability to achieve high sample quality and good mode coverage.
However, generating new samples is computationally expensive due to the need to simulate stochastic differential equations. 
Likelihood-based \glspl{GM} are complementary, in that they achieve lower sample quality, but sampling requires one model evaluation per sample and it is therefore extremely fast. %
While some attempts have been made to bridge the gap by combining \glspl{GAN} with \glspl{DM} \citep{XiaoICLR2022} or training \glspl{GAN} with diffusions \citep{WangICLR2023}, these still require careful engineering of architectures and training schedules. 
The observation that all \glspl{GM} share a common underlying objective indicates that we should look at what makes \glspl{DM} successful at optimizing their objective.
Then, the question we address in this paper is: can we borrow the strengths of score-based \glspl{DM} to improve likelihood-based \glspl{GM}, without paying the price of costly sample generation?

One distinctive element of score-based \glspl{DM} is data mollification, which is typically achieved by adding Gaussian noise~\citep{Song2019} or, in the context of image data sets, by blurring~\citep{Rissanen2023generative}.
A large body of evidence points to the {\em manifold hypothesis}~\citep{Roweis2000nonlinear}, which states that the intrinsic dimensionality of image data sets is much lower than the dimensionality of their input.
Density estimation in this context is particularly difficult because of the degeneracy of the likelihood for any density concentrated on the manifold where data lies~\citep{LoaizaGanem2022diagnosing}.
Under the manifold hypothesis, or even when the target density is multi-modal, the Lipschitz constant of \glspl{GM} has to be large, but regularization, which is necessary for robustness, is antagonist to this objective~\citep{SalmonaNeurIPS2022,Cornish20a}.
As we will study in detail in this paper, the process of data mollification gracefully guides the optimization mitigating manifold overfitting and enabling a desirable tail behavior, yielding accurate density estimation in low-density regions.
{
  In likelihood-based \glspl{GM}, data mollification corresponds to some form of simplification of the optimization objective.
  This type of approach, where the level of data mollification is annealed throughout training, can be seen as a continuation method \cite{Witkin87,Mohabi15}, which is a popular technique in the optimization literature to reach better optima.
}

  Strictly speaking, data mollification in score-based \glspl{DM} and likelihood-based \glspl{GM} are slightly different.
  In the latter, the amount of noise injected in the data is continuously annealed throughout training.
  At the beginning, the equivalent loss landscape seen by the optimizer is much smoother, due to the heavy perturbation of the data, and a continuous reduction of the noise level allows optimization to be gracefully guided until the point where the level of noise is zero~\cite{Witkin87,Mohabi15}.
  \glspl{DM}, instead, are trained at each step of the optimization process by considering \textbf{all} noise levels simultaneously, where complex amortization procedures, such as self-attention~\cite{Song2021scorebased}, allow the model to efficiently share parameters across different perturbation levels.
  It is also worth mentioning that score-based \glspl{DM} possess another distinctive feature in that they perform gradient-based density estimation \citep{Song2019,Hyvarinen05a}.  
  It has been conjectured that this can be helpful to avoid manifold overfitting by allowing for the modeling of complex densities while keeping the Lipschitz constant of score networks low~\citep{SalmonaNeurIPS2022}. %
  In this work, we attempt to verify the hypothesis that data mollification is heavily responsible for the success of score-based \glspl{DM}.
  We do so by proposing data mollification for likelihood-based \glspl{GM}, and provide theoretical arguments and experimental evidence that data mollification consistently improves their optimization.
Crucially, this strategy yields better sample quality and it is extremely easy to implement, as it requires adding very little code to any existing optimization loop. %

We consider a large set of experiments involving \glspl{VAE} and \glspl{NF} and some popular image data sets.
These provide a challenging test for likelihood-based \glspl{GM} due to the large dimensionality of the input space and to the fact that density estimation needs to deal with data lying on manifolds.
The results show systematic, and in some cases dramatic, improvements in sample quality, indicating that this is a simple and effective strategy to improve optimization of likelihood-based \glspl{GM} models.
The paper is organized as follows: 
in \cref{sec:methods} we illustrate the challenges associated with generative modeling when data points lie on a manifold, particularly with density estimation in low-density regions and manifold overfitting;
in \cref{sec:gen_mollification} we propose data mollification to address these challenges;
\cref{sec:experiments} reports the experiments with a discussion of the limitations and the broader impact, while \cref{sec:related} presents related works, and \cref{sec:conclusions} concludes the paper.

\section{Challenges in Training Deep Generative Models} \label{sec:methods}

We are interested in unsupervised learning, and in particular on the task of density estimation.
Given a dataset $\mathcal{D}$ consisting of $N$ \iid samples $\cD \stackrel{\Delta}{=} \{ \mbx_{i} \}_{i=1}^{N}$ with $\mbx_{i} \in \mathbb{R}^{D}$, we aim to estimate the unknown continuous generating distribution $p_{\text{data}}(\mbx)$.
In order to do so, we introduce a model $p_{\mbtheta}(\mbx)$ with parameters $\mbtheta$ and attempt to estimate $\mbtheta$ based on the dataset $\mathcal{D}$.
A common approach to estimate $\mbtheta$ is to maximize the likelihood  of the data, which is equivalent to minimizing the following objective:
\begin{align}
    \cL(\mbtheta) &\stackrel{\Delta}{=}   - \mathbb{E}_{p_{\text{data}}(\mbx)} \left[ \log p_{\mbtheta}(\mbx) \right] \label{eq:obj_likelihood}. %
\end{align}
There are several approaches to parameterize the generative model $p_{\mbtheta}(\mbx)$.
In this work, we focus on two widely used likelihood-based \glsreset{GM}\glspl{GM}, which are \glsreset{NF}\glspl{NF} \citep{Papamakarios2021, Kobyzev2021} and \glsreset{VAE}\glspl{VAE} \citep{KingmaW13,Rezende2015}.
Although \glspl{NF} and \glspl{VAE} are among the most popular deep \glspl{GM}, they are characterized by a lower sample quality compared to \glspl{GAN} and score-based \glspl{DM}.
In this section, we present two major reasons behind this issue by relying on the manifold hypothesis.

\subsection{The Manifold Hypothesis and Density Estimation in Low-Density Regions} %
  \label{sec:manifold_hypothesis}
The manifold hypothesis is a fundamental concept in manifold learning \citep{Roweis2000nonlinear,Tenenbaum2000global,Bengio2012RepresentationLA} stating that real-world high-dimensional data tend to lie on a manifold $\cM$ characterized by a much lower dimensionality compared to the one of the input space (ambient dimensionality) \citep{Narayanan2010}.
This has been verified theoretically and empirically for many applications and datasets \citep{Ozakin2009,Narayanan2010,Pope2021,Tempczyk2022}.
For example, \cite{Pope2021} report extensive evidence that natural image datasets have  indeed very low intrinsic dimension relative to the high number of pixels in the images.

\begin{wrapfigure}[16]{r}{0.42\textwidth}
    \vspace{-2ex}
    \centering
    \begin{subfigure}[b]{0.2\textwidth}
        \centering
        \caption*{\footnotesize Data distribution}
        \vspace{-1ex}
        \includegraphics[width=\textwidth]{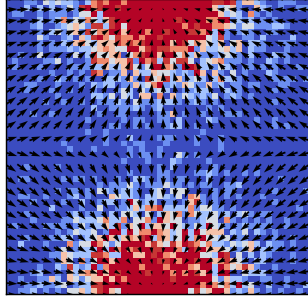}
    \end{subfigure}
    \begin{subfigure}[b]{0.2\textwidth}
        \centering
        \caption*{\footnotesize Estimated distribution}
        \vspace{-1ex}
        \includegraphics[width=\textwidth]{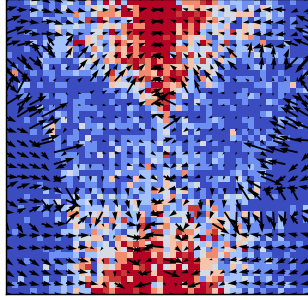}
    \end{subfigure}
    \caption{
        \textbf{Left:} Histogram of samples from data distribution $p_{\text{data}}(\mbx)$ and its true scores $\nabla_{\mbx} \log p_{\text{data}}(\mbx)$;
        \textbf{Right:} Histogram of of samples from the estimated distribution $p_{\mbtheta}(\mbx)$ and its scores $\nabla_{\mbx} \log p_{\mbtheta}(\mbx)$.
        In the low density regions, the model is unable to capture the true density and scores.
        \label{fig:toy_data}
    }
    \label{fig:1}
  \end{wrapfigure}

The manifold hypothesis suggests that density estimation in the input space is challenging and ill-posed.
In particular, data points on the manifold should be associated with high density, while points outside the manifold should be considered as lying in regions of nearly zero density \citep{Meng2021}.
This implies that the target density in the input space should be characterized by high Lipschitz constants. %
The fact that data is scarce in regions of low density makes it difficult to expect that models can yield accurate density estimation around the tails.
These pose significant challenges for the training of deep \glspl{GM} \citep{Cornish20a,Meng2021,Song2019}.
Recently, diffusion models \citep{Song2019,Ho2020,Song2021scorebased} have demonstrated the ability to mitigate this problem by gradually transforming a Gaussian distribution, whose support spans the full input space, into the data distribution.
This observation induces us to hypothesize that the data mollification mechanism in score-based \glspl{DM} is responsible for superior density estimation in low-density regions. 

To demonstrate the challenges associated with accurate estimation in low-density regions, we consider a toy experiment where we use a \textsc{real-nvp} flow \cite{Dinh2017density} to model a two-dimensional mixture of Gaussians, which is a difficult test for \glspl{NF} in general.
Details on this experiment are provided in the \cref{appendix:experiment_details}. %
\cref{fig:toy_data} depicts the true and estimated densities, and their corresponding scores, which are the gradient of the log-density function with respect to the data \citep{Hyvarinen05a}.
Note that the use of ``score'' here is slightly different from that from traditional statistics where score usually refers to the gradient of the log-likelihood with respect to model parameters.
As it can be seen in the figure, in regions of low data density, $p_{\mbtheta}(\mbx)$ is completely unable to model the true density and scores.
This problem is due to the lack of data samples in these regions and may be more problematic under the manifold hypothesis and for high-dimensional data such as images.
In \cref{sec:experiments}, we will demonstrate how it is possible to considerably mitigate this issue by means of data mollification.

\subsection{Manifold Overfitting}

The manifold hypothesis suggests that overfitting on a manifold can occur when the model $p_{\mbtheta}(\mbx)$ assigns an arbitrarily large likelihood in the vicinity of the manifold, even if the distribution does not accurately capture the true distribution $p_{\text{data}}(\mbx)$ \citep{Dai2018diagnosing,LoaizaGanem2022diagnosing}. 
{This issue is illustrated in Fig.~2 of \citep{LoaizaGanem2022diagnosing} and it will be highlighted in our experiment (\cref{sec:von_mises}), where the true data distribution $p_{\text{data}}(\mbx)$ is supported on a one-dimensional curve manifold $\cM$ in two-dimensional space $\mathbb{R}^{2}$.} 
Even when the model distribution $p_{\mbtheta}(\mbx)$ poorly approximates $p_{\text{data}}(\mbx)$, it may reach a high likelihood value by concentrating the density around the correct manifold $\cM$.
If $p_{\mbtheta}(\mbx)$ is flexible enough, any density defined on $\cM$ may achieve infinite likelihood and this might be an obstacle for retrieving $p_{\text{data}}(\mbx)$. %

A theoretical formalization of the problem of manifold overfitting appears in \cite{LoaizaGanem2022diagnosing} and it is based on the concept of Riemannian measure \citep{Pennec2006intrinsic}. 
The Riemannian measure on manifolds holds an analogous role to that of the Lebesgue measure on Euclidean spaces. 
To begin, we establish the concept of smoothness for a probability measure on a manifold.

\begin{definition}
    Let $\mathcal{M}$ be a finite-dimensional manifold, and $p$ be a probability measure on $\mathcal{M}$.
    Let $\mathfrak{g}$ be a Riemannian metric on $\mathcal{M}$ and $\mu_\mathcal{M}^{(\mathfrak{g})}$ the corresponding Riemannian measure.
    We say that $p$ is \emph{smooth} if $p \ll \mu_\mathcal{M}^{(\mathfrak{g})}$ and it admits a continuous density $p:\mathcal{M} \rightarrow \mathbb{R}_{>0}$ with respect to $\mu_\mathcal{M}^{(\mathfrak{g})}$.
\end{definition}
We now report Theorem~1 from \cite{LoaizaGanem2022diagnosing} followed by a discussion on its implications for our work.
\begin{theorem}
    (Gabriel Loaiza-Ganem et al. \cite{LoaizaGanem2022diagnosing}). Let $\mathcal{M} \subset \mathbb{R}^{D}$ be an analytic $d$-dimensional embedded submanifold of $\mathbb{R}^d$ with $d < D$, $\mu_D$ is  the Lebesgue measure on $\mathbb{R}^{D}$, and $p^\dagger$ a smooth probability measure on $\mathcal{M}$.
    Then there exists a sequence of probability measures $\big\{p_{\mbtheta}^{(t)}\big\}_{t=0}^{\infty}$ on $\mathbb{R}^{D}$ such that:
    \vspace{-1.5ex}
    \begin{enumerate}
        \item $p_{\mbtheta}^{(t)} \rightarrow p^{\dagger}$ as $t \rightarrow \infty$.
        \vspace{-1ex}
        \item $\forall t \geq 0, p_{\mbtheta}^{(t)} \ll \mu_D$ and $p_{\mbtheta}^{(t)}$ admits a density $p_{\mbtheta}^{(t)} : \mathbb{R}^{D} \rightarrow \mathbb{R}_{>0}$ with respect to $\mu_D$ such that:
        \vspace{-1ex}
        \begin{enumerate}
            \item $\lim_{t\rightarrow \infty} p_{\mbtheta}^{(t)}(\mbx) = \infty$, $\forall \mbx \in \mathcal{M}$.
            \vspace{-0.5ex}
            \item $\lim_{t\rightarrow \infty} p_{\mbtheta}^{(t)}(\mbx) = 0$, $\forall \mbx \notin \text{cl}(\mathcal{M})$, where $\text{cl}(\cdot)$  denotes closure in $\mathbb{R}^{D}$.
        \end{enumerate}
    \end{enumerate}
    \label{thm:manifold_overfitting}
\end{theorem}
\cref{thm:manifold_overfitting} holds for any smooth probability measure supported in $\mathcal{M}$.
This is an important point because this includes the desired $p_{\text{data}}$, provided that this is smooth too.
The key message in~\cite{LoaizaGanem2022diagnosing} is that, a-priori, there is no reason to expect that for a likelihood-based model to converge to $p_{\text{data}}$ out of all the possible $p^{\dagger}$.
Their proof is based on convolving $p^{\dagger}$ with a Gaussian kernel with variance $\sigma_t^2$ that decreases to zero as $t \rightarrow \infty$, and then verify that the stated properties of $p_{\mbtheta}^{(t)}$ hold.
Our analysis, while relying on the same technical tools, is instead constructive in explaining why the proposed data mollification allows us to avoid manifold overfitting.
The idea is as follows: at time step $t = 0$, we select the desired $p_{\text{data}}$ convolved with a Gaussian kernel with a large, but finite, variance $\sigma^2(0)$ as the target distribution for the optimization.
Optimization is performed and $p_{\mbtheta}^{(0)}$ targets %
this distribution, without any manifold overfitting issues, since the dimensionality of the corrupted data is non-degenerate.
At the second step, the target distribution is obtained by convolving $p_{\text{data}}$ with the kernel with variance $\sigma^2(1) < \sigma^2(0)$, and again manifold overfitting is avoided.
By iteratively repeating this procedure, we can reach the point where we are matching a distribution convolved with an arbitrarily small variance $\sigma^2(t)$, without ever experiencing manifold overfitting.
{
  When removing the last bit of perturbation we fall back to the case where we experience manifold overfitting.
  However, when we operate in a stochastic setting, which is the typical scenario for the \glspl{GM} considered here, we avoid ending up in solutions for which the density is degenerate and with support which is exactly the data manifold.
  Another way to avoid instabilities is to adopt gradient clipping.
  Note that, as mentioned in~\cite{LoaizaGanem2022diagnosing} and verified by ourselves in earlier investigations, a small constant amount of noise does not provide any particular benefits over the original scheme, whereas gradually reducing the level of data mollification improves optimization dramatically.

}

{
  \subsection{Data Mollification as a Continuation Method}
  We can view the proposed data mollification approach as a continuation method \cite{Witkin87,Mohabi15}.
  Starting from the target objective function, which in our case is $\cL(\mbtheta)$ in \cref{eq:obj_likelihood} (or a lower bound in the case of \glspl{VAE}), we construct a family of functions $\cH(\mbtheta, \gamma)$ parameterized by an auxiliary variable $\gamma \in [0, 1]$ so that $\cH(\mbtheta, 0) = \cL(\mbtheta)$.
  The objective functions $\cH(\mbtheta, \gamma)$ are defined so that the higher $\gamma$ the easier is to perform optimization.
  In our case, when $\gamma = 1$ we operate under a simple regime where we target a Gaussian distribution, and likelihood-based \glspl{GM} can model these rather easily. 
  By annealing $\gamma$ from $1$ to $0$ with a given schedule, the sequence of optimization problems with objective $\cH(\mbtheta, \gamma)$ is increasinly more complex to the point where we target $\cL(\mbtheta)$.
  In essence, the proposed data mollification approach can be seen as a good initialization method, as the annealing procedure introduces a memory effect in the optimization process, which is beneficial in order to obtain better optima.
}

\section{Generative Models with Data Mollification} \label{sec:gen_mollification}

Motivated by the aforementioned problems with density estimation in low-density regions and manifold overfitting, we propose a simple yet effective approach to improve likelihood-based \glspl{GM}.
Our method involves mollifying data using Gaussian noise, gradually reducing its variance, until recovering the original data distribution $p_{\text{data}}(\mbx)$.
This mollification procedure is similar to the reverse process of diffusion models, where a prior noise distribution is smoothly transformed into the data distribution \citep{Song2019, Ho2020, Song2021scorebased}.
As already mentioned, data mollification alleviates the problem of manifold overfitting and it induces a memory effect in the optimization which improves density estimation in regions of low density.

\begin{figure}[H]
    \begin{minipage}{0.455 \textwidth}
        \centering
        \includegraphics[width=0.99\textwidth]{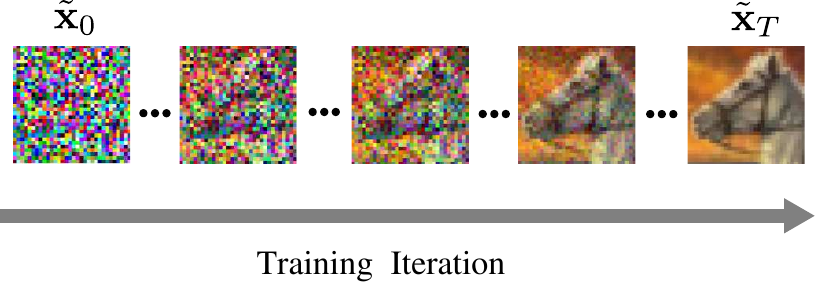}

        \vspace{-0.8ex}

        \caption{Illustration of Gaussian mollification, where $\tilde{\mbx}_t$ is the mollified data at iteration $t$.
    \label{fig:mollification}
    }
    \end{minipage}
    \hfill
    \begin{minipage}{0.51\textwidth}
        \scalebox{.85}{
            \begin{algorithm}[H]
                \caption{Gaussian mollification} \label{alg:gaussian_mollification}
                \For{$t \leftarrow 1, 2, ..., T$}{
                    $\mbx \sim p_{\text{data}}(\mbx)$ \tcp{\footnotesize{Sample training data}}
                    {\color{BrickRed}{$\tilde{\mbx}_t = \alpha_t \mbx + \sigma_t \boldsymbol{\varepsilon}$}} \tcp{\footnotesize{{\textbf{Mollify data with $\alpha_t, \sigma^{2}_t \leftarrow \gamma(t/T)$ and $\boldsymbol{\varepsilon} \sim \cN(\boldsymbol{0}, \mbI)$}}}}
                    $\mbtheta_t \leftarrow \textsc{update}(\mbtheta_{t-1}, \tilde{\mbx}_t)$ \tcp{\footnotesize{Train model}}
                }
            \end{algorithm}
        }
    \end{minipage}
    
\end{figure}

\paragraph{Gaussian Mollification.}

Given that we train the model $p_{\mbtheta}(\mbx)$ for $T$ iterations, we can create a sequence of progressively less smoothed versions of the original data $\mbx$, which we refer to as mollified data $\tilde{\mbx}_t$.
Here, $t$ ranges from $t=0$ (the most mollified) to $t=T$ (the least mollified). 
For any $t \in [0,T]$, the distribution of the mollified data $\tilde{\mbx}_t$, conditioned on $\mbx$, is given as follows:
\begin{equation}
    q(\tilde{\mbx}_{t} \g \mbx) = \mathcal{N}(\tilde{\mbx}_{t}; \alpha_t \mbx, \sigma_t^2 \mbI),
\end{equation}
where $\alpha_t$ and $\sigma^2_t$ are are positive scalar-valued functions of $t$. In addition, we define the signal-to-noise ratio %
$
\text{SNR}(t) = {\alpha_t^2} / {\sigma_t^2}.
$
and we assume that it monotonically increases with $t$, i.e., $\text{SNR}(t) \leq \text{SNR}(t+1)$ for all $t \in [0,T-1]$.
In other words, the mollified data $\tilde{\mbx}_t$ is progressively less smoothed as $t$ increases.
In this work, we adopt the \textit{variance-preserving} formulation used for diffusion models \citep{SohlDickstein15,Ho2020,Kingma2021}, where $\alpha_t = \sqrt{1 - \sigma_t^2}$ and $\sigma^{2}_t = \gamma(t/T)$.
Here, $\gamma(\cdot)$ is a monotonically decreasing function from $1$ to $0$ that controls the rate of mollification.
Intuitively, this procedure involves gradually transforming an identity-covariance Gaussian distribution into the distribution of the data.
\cref{alg:gaussian_mollification} summarizes the proposed Gaussian mollification procedure, where the {\color{BrickRed} red line} indicates a simple additional step required to mollify data compared with vanilla training using the true data distribution.

\vspace{-1ex}

\paragraph{Noise schedule.} 
The choice of the noise schedule $\gamma(\cdot)$ has an impact on the performance of the final model.
In this work, we follow common practice in designing the noise schedule based on the literature of score-based \glspl{DM} \citep{Nichol21a,Hoogeboom2023simple,Chen2023}.
In particular, we adopt a sigmoid schedule \citep{Jabri2022}, which has recently been shown to be more effective in practice compared to other choices such as linear \citep{Ho2020} or cosine schedules \citep{Nichol21a}.
\begin{figure}[H]
    \begin{minipage}{0.53 \textwidth}
        The sigmoid schedule $ \gamma(t/T)$ \citep{Jabri2022} is defined through the sigmoid function: %
        \begin{equation}
         \text{sigmoid}\left(- \frac{t/T }{\tau}  \right), 
        \end{equation}
        where $\tau$ is a temperature hyper-parameter. This function is then scaled and shifted to ensure that $\gamma(0)=1$ and $\gamma(1)=0$. %
        We encourage the reader to check the implementation of this schedule, available in \cref{appendix:noise_schedules}.
        \cref{fig:sigmoid_schedule} illustrates the sigmoid schedule and the corresponding $\log(\text{SNR})$ with different values of $\tau$.
    We use a default temperature of $0.7$ as it demonstrates consistently good results in our experiments.
    \end{minipage} \hfill
    \begin{minipage}{0.45 \textwidth}
        \tikzexternaldisable
        \scriptsize
        \centering
        \setlength{\figurewidth}{3.8cm}
        \setlength{\figureheight}{3.4cm}
        \input{figures/sigmoid_schedule.tikz}
        \tikzexternalenable
        \vspace{-4ex}
        \caption{Illustration of sigmoid schedule and the corresponding $\log(\text{SNR})$.
        The temperature values from $0.2$ to $0.9$ are progressively shaded, with the lighter shade corresponding to lower temperatures.
        } \label{fig:sigmoid_schedule}
    \end{minipage}
\end{figure}

\vspace{-3ex}

\section{Experiments} \label{sec:experiments}

In this section, we demonstrate emprically the effectiveness of our proposal through a wide range of experiments on synthetic data, and some popular real-world tabular and image data sets.
\cref{appendix:experiment_details} contains a detailed description of each experiment to guarantee reproducibility.

\subsection{2D Synthetic Data Sets}

We begin our experimental campaign with two 2D synthetic data sets.
The two-dimensional nature of these data sets allows us to demonstrate the effectiveness of Gaussian mollification in mitigating the challenges associated with density estimation in low-density regions and manifold overfitting. 
Here, we consider $p_{\mbtheta}(\mbx)$ to be a \textsc{real-nvp} flow \citep{Dinh2017density}, which comprises five coupling bijections, each consisting of a two-hidden layer \gls{MLP}.
To assess the capability of $p_{\mbtheta}(\mbx)$ to recover the true data distribution, we use \gls{MMD} \citep{Gretton12} with a \gls{RBF} kernel on a held-out set.
In these experiments, we employ the Gaussian mollification strategy presented in the previous section and compare the estimated density with the \textit{vanilla} approach where we use the original training data without any mollification.

\begin{figure}[t]
    \begin{minipage}{0.73 \textwidth}
        \centering
        \includegraphics[width=\textwidth]{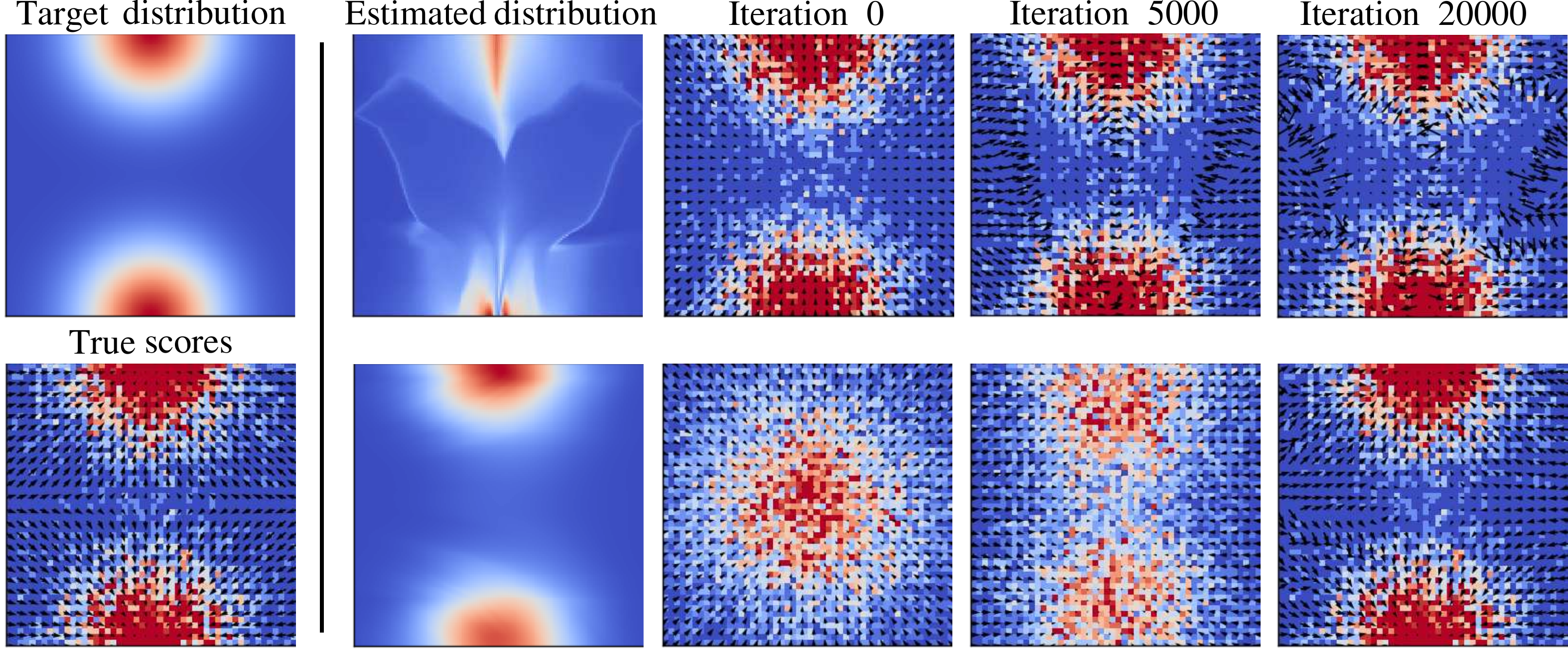}
        \caption{The first column shows the target distribution and the true scores.
            The second column depicts the estimated distributions of the \gls{GMM} \label{fig:gmm_results}, which yield ${\scriptsize\text{MMD}^2}$ of ${\scriptsize 15.5}$ and ${\scriptsize 2.5}$ for the vanilla (top) and mollification (bottom) training, respectively.
            The remaining columns show histogram of samples from the true (\textbf{top row}) and mollified data (\textbf{bottom row}), and estimated scores.
            \label{fig:gmm_scores}
        }
    \end{minipage} \hfill
    \begin{minipage}{0.25 \textwidth}
        \tikzexternaldisable
        \scriptsize
        \centering
        \setlength{\figurewidth}{4.2cm}
        \setlength{\figureheight}{3.3cm}
        \input{figures/toy/gmm.tikz}

        \vspace{-2ex}
        
        \definecolor{color_1}{HTML}{427fb1}
\definecolor{color_2}{HTML}{b62428}
\tikzexternaldisable
{\setlength{\tabcolsep}{1.8pt}
    \scriptsize
    \begin{tabular}{clcl}
        {\protect\tikz[baseline=-1ex]\protect\draw[color=color_1, fill=color_1, opacity=1, mark size=1.7pt, line width=1.7pt] plot[] (-0.0,0)--(-0.35,0);} & \textsf{Vanilla} &
        {\protect\tikz[baseline=-1ex]\protect\draw[color=color_2, fill=color_2,  opacity=1, mark size=1.7pt, line width=1.7pt] plot[] (-0.0,0)--(-0.35,0);} &  \textsf{Mollification} \\
    \end{tabular}\tikzexternalenable
}
        \tikzexternalenable

        \vspace{-1ex}
        \caption{The learning curves of the \gls{GMM} experiments.  \label{fig:gmm_curves}}
    \end{minipage}
\end{figure}

\begin{figure}[b]
    \centering
    \includegraphics[width=\textwidth]{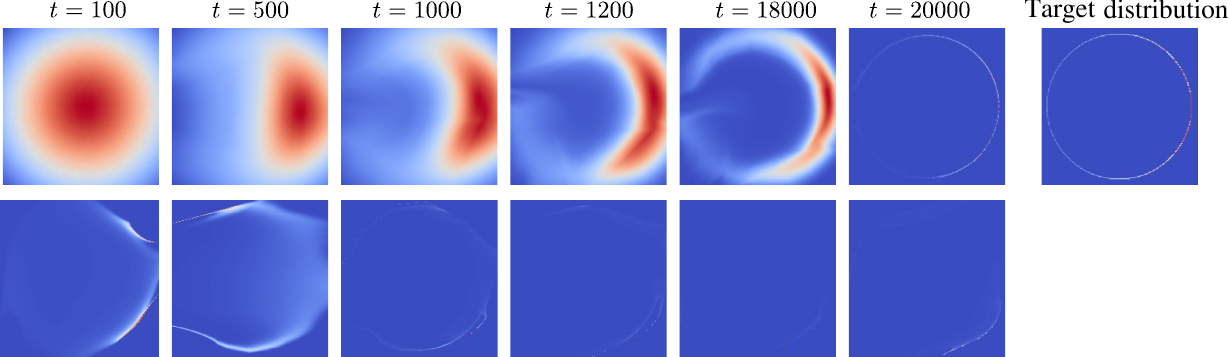}
    \caption{The progression of the estimated densities for the von Mises distribution from the vanilla (\textbf{bottom row}) and our mollification (\textbf{top row}) approaches.
    \label{fig:von_mises}
    }
\end{figure}

\paragraph{Mixture of Gaussians.}
First, we consider a target distribution that is a mixture of two Gaussians, as depicted in \cref{fig:gmm_scores}. 
As discussed in \cref{sec:manifold_hypothesis}, the vanilla training procedure fails to accurately estimate the true data distribution and scores, particularly in the low-density regions.
The estimated densities and the mollified data during the training are depicted in \cref{fig:gmm_scores}.
Initially, the mollification process considers a simpler coarse-grained version of the target density, which %
is easy to model.
This is demonstrated by the low training loss at the beginning of the optimization, as depicted in \cref{fig:gmm_curves}.
Subsequently, the method gradually reduces the level of noise allowing for a progressive refinement of the estimated versions of the target density.
This process uses the solution from one level of mollification as a means to guiding optimization for the next.
As a result, Gaussian mollification facilitates the recovery of the modes and enables effective density estimation in low-density regions.
The vanilla training procedure, instead, produces a poor estimate of the target density, as evidenced by the trace-plot of the $\gls{MMD}^2$ metric in \cref{fig:gmm_curves} and the visualization of the scores in \cref{fig:gmm_scores}.

\paragraph{Von Mises distribution.} \label{sec:von_mises}

We proceed with an investigation of the von Mises distribution on the unit circle, as depicted in \cref{fig:von_mises}, with the aim of highlighting the issue of manifold overfitting \citep{LoaizaGanem2022diagnosing}. 
In this experiment, the data lies on a one-dimensional manifold embedded in a two-dimensional space. 
The vanilla training procedure fails to approximate the target density effectively, as evidenced by the qualitative results and the substantially high value of $\gls{MMD}^2$ ({\small $\approx 383.44$}) shown in \cref{fig:von_mises}. 
In contrast, Gaussian mollification gradually guides the estimated density towards the target, as depicted in \cref{fig:von_mises}, leading to a significantly lower $\gls{MMD}^2$ ({\small $\approx 6.13$}). 
Additionally, the mollification approach enables the estimated model not only to accurately learn the manifold but also to capture the mode of the density correctly.

\subsection{Image Experiments}

\begin{table}[H]
    \caption{\textsc{fid} score on \cifar and \celeba dataset  (\emph{lower is better}).
        The small colored numbers indicate {\color{positive_green} improvement} or {\color{negative_red} degration} of the mollification training compared to the vanilla training.
    }
    \vspace{1ex}

    \centering

    \resizebox{0.99\columnwidth}{!}{
        \setlength{\tabcolsep}{5pt}
        \setlength\extrarowheight{0.5pt}
        \begin{sc}
            \small
            \rowcolors{2}{}{mylightgray}

            \begin{tabular}{r||ccc|ccc}
                \toprule
                \multicolumn{1}{c||}{\multirow{2}{*}{Model}} & \multicolumn{3}{c|}{cifar10} & \multicolumn{3}{c}{celeba}                                                                                                                                                                                                                                                                    \\ [1.9pt] \cline{2-7}
                \multicolumn{1}{c||}{}                       & \multicolumn{1}{c}{vanilla}  & \multicolumn{1}{c}{\begin{tabular}[c]{@{}c@{}}gaussian\\ mollifcation\end{tabular}}                  & \multicolumn{1}{c|}{\begin{tabular}[c]{@{}c@{}}blurring\\ mollification\end{tabular}}                 & \multicolumn{1}{c}{vanilla} & \multicolumn{1}{c}{\begin{tabular}[c]{@{}c@{}}gaussian\\ mollification\end{tabular}}                 & \multicolumn{1}{c}{\begin{tabular}[c]{@{}c@{}}blurring\\ mollificatoin\end{tabular}}                 \\
                \midrule
                \midrule
                real-nvp \citep{Dinh2017density}             & $131.15$                     & $121.75$ { \tiny \color{positive_green} $\downarrow 7.17$\% }  & $120.88$ { \tiny \color{positive_green} $\downarrow 7.83$\% }  & $81.25$                     & $79.68$ { \tiny \color{positive_green} $\downarrow 1.93$\% }  & $85.40$ { \tiny \color{negative_red} $\uparrow 5.11$\% }      \\
                glow \citep{Kingma2018Glow}                  & $74.62$                      & $64.87$ { \tiny \color{positive_green} $\downarrow 13.07$\% }  & $66.70$ { \tiny \color{positive_green} $\downarrow 10.61$\% }  & $97.59$                     & $70.91$ { \tiny \color{positive_green} $\downarrow 27.34$\% } & $74.74$ { \tiny \color{positive_green} $\downarrow 23.41$\% } \\
                \midrule
                vae\citep{KingmaW13}                         & $191.98$                     & $155.13$ { \tiny \color{positive_green} $\downarrow 19.19$\% } & $175.40$ { \tiny \color{positive_green} $\downarrow 8.64$\% }   & $80.19$                     & $72.97$ { \tiny \color{positive_green} $\downarrow 9.00$\% }  & $77.29$ { \tiny \color{positive_green} $\downarrow 3.62$\% }  \\
                vae-iaf \citep{Kingma2016}                   & $193.58$                     & $156.39$ { \tiny \color{positive_green} $\downarrow 19.21$\% } & $162.27$ { \tiny \color{positive_green} $\downarrow 16.17$\% } & $80.34$                     & $73.56$ { \tiny \color{positive_green} $\downarrow 8.44$\% }  & $75.67$ { \tiny \color{positive_green} $\downarrow 5.81$\% }  \\
                iwae \citep{Burda2015importance}             & $183.04$                     & $146.70$ { \tiny \color{positive_green} $\downarrow 19.85$\% } & $163.79$ { \tiny \color{positive_green} $\downarrow 10.52$\% } & $78.25$                     & $71.38$ { \tiny \color{positive_green} $\downarrow 8.78$\% }  & $76.45$ { \tiny \color{positive_green} $\downarrow 2.30$\% }  \\
                $\beta$-vae \citep{Higgins2017betavae}       & $112.42$                     & $93.90$ { \tiny \color{positive_green} $\downarrow 16.47$\% }  & $101.30$ { \tiny \color{positive_green} $\downarrow 9.89$\% } & $67.78$                     & $64.59$ { \tiny \color{positive_green} $\downarrow 4.71$\% }  & $67.08$ { \tiny \color{positive_green} $\downarrow 1.03$\% }  \\
                hvae \citep{Caterini2018}                     & $172.47$                     & $137.84$ { \tiny \color{positive_green} $\downarrow 20.08$\% } & $147.15$ { \tiny \color{positive_green} $\downarrow 14.68$\% } & $74.10$                     & $72.28$ { \tiny \color{positive_green} $\downarrow 2.46$\% }   & $77.54$ { \tiny \color{negative_red} $\uparrow 4.64$\% }      \\
                \bottomrule
            \end{tabular}
        \end{sc}
    }
    \label{tab:fid}
\end{table}

\paragraph{Setup.}
We evaluate our method on image generation tasks on \cifar \citep{Krizhevsky2009learning} and \celeba $64$ \citep{LiuLWT15} datasets, using a diverse set of likelihood-based \glspl{GM}.
The evaluated models include the vanilla {\textsc{vae}} \cite{KingmaW13}, the {\textsc{${\beta}$-vae}} \citep{Higgins2017betavae}, and the {\textsc{vae-iaf}} \citep{Kingma2016} which employs an expressive inverse autoregressive flow for the approximate posterior.
To further obtain flexible approximations of the posterior of latent variables as well as a tight \gls{ELBO}, we also select the {Hamiltonian-\textsc{vae} (\textsc{hvae})} \citep{Caterini2018} and the {importance weighted \textsc{vae} (\textsc{iwae})} \citep{Burda2015importance}.
For flow-based models, we consider the \textsc{real-nvp} \citep{Dinh2017density} and \textsc{glow} \citep{Kingma2018Glow} models in our benchmark.
We found that further training the model on the original data after the mollification procedure leads to better performance. 
Hence, in our approach we apply data mollification during the first half of the optimization phase, and we continue optimize the model using the original data in the second half.
Nevertheless, to ensure a fair comparison, we adopt identical settings for the vanilla and the proposed approaches, including random seed, optimizer, and the total number of iterations.

\paragraph{Blurring mollification.}
Even though Gaussian mollification is motivated by the manifold hypothesis, it is not the only way to mollify the data.
Indeed, Gaussian mollification does not take into account certain inductive biases that are inherent in natural images, including their multi-scale nature. 
Recently, \cite{Rissanen2023generative,Hoogeboom2023blurring,Daras2022soft} have proposed methods that incorporate these biases in diffusion-type generative models. 
Their approach involves stochastically reversing the heat equation, which is a \gls{PDE} that can be used to erase fine-scale information when applied locally to the 2D plane of an image.
During training, the model first learns the coarse-scale structure of the data, which is easier to learn, and then gradually learns the finer details.
It is therefore interesting to assess whether this form of data mollification is effective in the context of this work compared to the addition of Gaussian noise.
Note, however, that under the manifold hypothesis, this type of mollification produces the opposite effect to the addition of Gaussian noise in that at time $t=0$ mollified images lie on a 1D manifold and they are gradually transformed to span the dimension of the data manifold; more details on blurring mollification can be found in \cref{appendix:blur_mollification}.

\begin{figure}[H]
    \hfill
    \begin{minipage}{0.3\textwidth}
        \centering

        \tikzexternaldisable
        \scriptsize
        \centering
        \setlength{\figurewidth}{4.5cm}
        \setlength{\figureheight}{3.2cm}
        \begin{tikzpicture}

\definecolor{darkgray176}{RGB}{176,176,176}
\definecolor{firebrick}{RGB}{178,34,34}
\definecolor{gainsboro}{RGB}{220,220,220}
\definecolor{seagreen}{RGB}{46,139,87}
\definecolor{steelblue}{RGB}{70,130,180}

\begin{axis}[
height=\figureheight,
major tick length=1ex,
tick align=outside,
tick pos=left,
width=\figurewidth,
x grid style={darkgray176},
xlabel={Epoch},
xmin=0, xmax=205,
xtick style={color=black},
y grid style={darkgray176},
ylabel={\textsc{vae} - FID [log]},
ymode=log,
ymin=130, ymax=400,
ytick style={color=black}
]

\addplot [very thick, steelblue]
table {%
9 275.828360458979
19 258.475466804566
29 247.27933457069
39 243.317551677499
49 232.135458295369
59 222.347000737537
69 211.870721571321
79 205.543648883099
89 211.264828828225
99 205.192069721024
109 203.00336635333
119 201.409334983278
129 197.598290246665
139 194.017035167594
149 194.559061784145
159 191.942091427875
169 193.374292980827
179 190.477099855304
189 192.669358361209
200 191.986595275903
};
\addplot [very thick, firebrick]
table {%
9 284.995885067403
19 203.668304601768
29 185.739520570731
39 181.499296673885
49 183.763396123353
59 180.733472815963
69 173.470436966033
79 169.069473691144
89 169.708200350771
99 164.598705842794
109 163.932112087644
119 162.566692590737
129 160.904043429702
139 153.66002872769
149 159.640951948866
159 156.903666153181
169 154.67151302412
179 154.713318310392
189 153.000666174139
200 155.130112823793
};
\addplot [very thick, seagreen]
table {%
9 359.28524354459
19 383.134439210067
29 353.400773050816
39 376.956311211456
49 350.084101159933
59 298.390937790071
69 266.571382091969
79 239.090563455245
89 223.401217018332
99 209.097146174629
109 191.597088178865
119 187.980680791696
129 182.280596131415
139 176.236841885393
149 175.193717791509
159 171.662691253393
169 164.862513397538
179 165.17332998973
189 163.887022983427
200 162.695184159534
};
\draw (axis cs:0.9,0.9) node[
  scale=0.5,
  anchor=north west,
  text=black,
  rotate=0.0
]{VAE};
\end{axis}

\end{tikzpicture}
        
        \vspace{-2ex}
        
        \centering
        \hspace{-2ex}\begin{tikzpicture}

\definecolor{darkgray176}{RGB}{176,176,176}
\definecolor{firebrick}{RGB}{178,34,34}
\definecolor{seagreen}{RGB}{46,139,87}
\definecolor{steelblue}{RGB}{70,130,180}

\begin{axis}[
height=\figureheight,
major tick length=1ex,
tick align=outside,
tick pos=left,
width=\figurewidth,
x grid style={darkgray176},
xlabel={Epoch},
xmin=-2.95, xmax=83.95,
xtick style={color=black},
y grid style={darkgray176},
ylabel={\textsc{glow} - FID [log]},
ymode=log,
ymin=60, ymax=440,
ytick style={color=black}
]
\addplot [very thick, steelblue]
table {%
1 219.87905533115
2 173.191181029243
4 132.807374023949
5 122.562572364224
7 106.853497449081
9 102.342958953606
10 90.5040287142261
12 91.7662795858768
13 87.8931117376714
15 89.9819737912088
17 81.2961696439393
18 79.8161413690451
20 80.3415839962055
21 81.2312359009489
23 79.852354177318
25 80.1597453835761
26 78.7095287263228
28 76.8321190404832
30 78.9191622234901
31 78.2146108053648
33 80.0399382814894
34 80.5643798520238
36 78.6578736525902
38 78.9095321516281
39 76.8228420780823
41 77.7902512388748
42 78.5054324951778
44 75.5641108620846
46 76.3896550518472
47 76.1175611462584
49 79.6537578746575
50 76.4976338942581
52 77.9594513673208
54 75.5071758217979
55 76.2651715831182
57 76.7448804506168
59 75.8643348810895
60 75.1648242279502
62 77.3130080543195
63 75.9836639438986
65 75.698175954769
67 76.8550461015116
68 77.0208939817639
70 75.5197660572222
71 77.0376830954293
73 73.8980899895075
75 73.7943582061472
76 75.9738976074497
78 76.4807633695735
80 74.6266966989407
};
\addplot [very thick, firebrick]
table {%
1 434.259901707134
2 430.817218383184
4 420.210304277221
5 404.416296951276
7 395.01995870988
9 383.083867296336
10 368.712904477546
12 358.622953604526
13 348.183697283412
15 333.016386483374
17 328.098336521876
18 318.191993476283
20 307.647145662772
21 296.042305131146
23 284.969471971731
25 282.502110738499
26 276.041382367121
28 271.194811842369
30 261.071130497678
31 252.882721725255
33 249.074045275904
34 238.09344890853
36 232.048267991833
38 218.144833914201
39 213.114928036612
41 202.213959405462
42 187.270615480434
44 175.044414854995
46 148.535906321917
47 89.8168670518874
49 76.7439963468889
50 74.1820498508703
52 69.6880934836911
54 69.9143271479604
55 68.0051366495035
57 67.1408933526313
59 66.3872527106699
60 68.3246358922891
62 67.952158344012
63 66.7338470531034
65 66.3287869939228
67 68.5069074628458
68 66.6535026567948
70 67.0255801531943
71 65.6370019018968
73 64.8816345636447
75 66.0719062518673
76 64.830590898642
78 66.4892999039883
80 64.8788928186547
};
\addplot [very thick, seagreen]
table {%
1 306.806753698433
2 367.854737605121
4 354.609023435558
5 353.916827707974
7 399.726268439631
9 429.128394934453
10 412.110150386918
12 403.436322395884
13 395.361689967761
15 373.058248915857
17 348.00470100772
18 304.45703821484
20 273.595084231463
21 249.771735853673
23 229.551222448051
25 226.889186110185
26 214.485662528359
28 203.929066563399
30 194.391242739334
31 189.508004103682
33 185.930704040347
34 190.602166984679
36 176.566396554872
38 156.82674875017
39 149.524483606601
41 157.946103661143
42 126.31331121423
44 110.406804467264
46 104.532674661716
47 93.1481007848361
49 97.8761246065096
50 88.3844309070877
52 80.2218248330981
54 76.8936127838215
55 78.4675253367593
57 77.8098616388431
59 73.7843618506092
60 71.0150231408294
62 72.0647587974309
63 69.8286146112885
65 72.2446970756613
67 71.7477855029989
68 69.344385847102
70 69.1409830725254
71 68.0282119467219
73 69.7985872284308
75 72.2093027720322
76 68.3974138040853
78 67.1394089982266
80 67.2945465828857
};
\end{axis}

\end{tikzpicture}
        \tikzexternalenable

        \definecolor{color_1}{HTML}{427fb1}
\definecolor{color_2}{HTML}{b62428}
\definecolor{color_3}{HTML}{2e8b57}
\tikzexternaldisable
{\setlength{\tabcolsep}{1.8pt}
    \scriptsize
    \begin{tabular}{clclcl}
        {\protect\tikz[baseline=-1ex]\protect\draw[color=color_1, fill=color_1, opacity=1, mark size=1.7pt, line width=1.7pt] plot[] (-0.0,0)--(-0.15,0);} & \textsf{Vanilla} &
        {\protect\tikz[baseline=-1ex]\protect\draw[color=color_2, fill=color_2,  opacity=1, mark size=1.7pt, line width=1.7pt] plot[] (-0.0,0)--(-0.15,0);} &  \textsf{Gauss.} &
        {\protect\tikz[baseline=-1ex]\protect\draw[color=color_3, fill=color_3,  opacity=1, mark size=1.7pt, line width=1.7pt] plot[] (-0.0,0)--(-0.15,0);} &  \textsf{Blurring} \\
    \end{tabular}\tikzexternalenable
}

        \vspace{-1ex}

        \caption{The progression of FID on \cifar dataset.
            \label{fig:fid_curves}%
        }
    \end{minipage}
    \hfill
    \begin{minipage}{0.67 \textwidth}
            \centering
            \includegraphics[width=\textwidth]{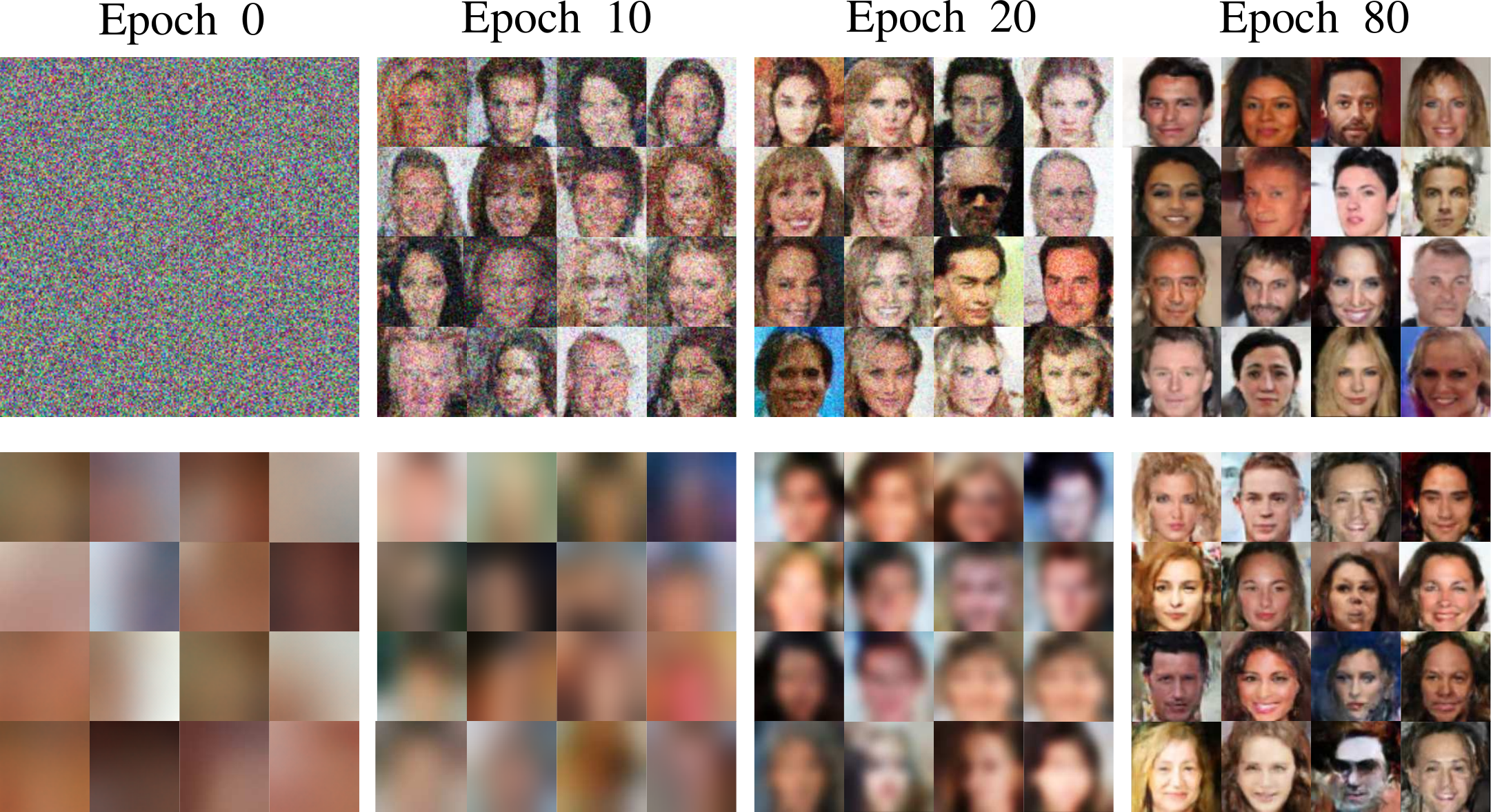}
            \caption{Intermediate samples generated from \textsc{real-nvp} flows \citep{Dinh2017density}, which are trained on \celeba dataset employed with Gaussian (\textbf{top row}) and blurring mollification (\textbf{bottom row}).
            \label{fig:celeba_figs}}
    \end{minipage}

\end{figure}

\paragraph{Image generation.}
We evaluate the quality of the generated images using the popular \gls{FID} score \citep{Heusel2017} computed on $50\mathrm{K}$ samples from the trained model using the \texttt{pytorch-fid} \footnote{\url{https://github.com/mseitzer/pytorch-fid}} library.
The results, reported in \cref{tab:fid}, indicate that the proposed data mollification consistently improves model performance compared to vanilla training across all datasets and models.
Additionally, mollification through blurring, which is in line with recent results from diffusion models \citep{Rissanen2023generative}, is less effective than Gaussian mollification, although it still enhances the vanilla training in most cases.
We also show intermediate samples in \cref{fig:celeba_figs} illustrating the progression of samples from pure random noise or {completely blurred images} %
to high-quality images. 
Furthermore, we observe that Gaussian mollification leads to faster convergence of the \textsc{fid} score for \gls{VAE}-based models, as shown in \cref{fig:fid_curves}. 
We provide additional results in \cref{appendix:additional_results}.
As a final experiment, we consider a recent large-scale \gls{VAE} model for the \cifar data set, which is a deep hierarchical \gls{VAE} (\textsc{nvae}) \citep{Vahdat2020}.
By applying Gaussian mollification without introducing any additional complexity, e.g., step-size annealing, we improve the \gls{FID} score from $53.64$ to $52.26$.

\begin{minipage}{0.55\textwidth}
    \paragraph{Choice of noise schedule.}
    We ablate on the choice of noise schedule for Gaussian mollification.
    Along with the sigmoid schedule, we also consider the linear \citep{Ho2020} and cosine \citep{Nichol21a} schedules, which are also popular for diffusion models. 
    As shown in Table \ref{tab:noise_schedule}, our method consistently outperforms the vanilla approach under all noise schedules.
    We also observe that the sigmoid schedule consistently produced good results.
    Therefore, we chose to use the sigmoid schedule in all our experiments.
\end{minipage}\hfill
\begin{minipage}{0.43\textwidth}
    \vspace{-5.8ex}
    \begin{table}[H]
    \caption{ \textsc{fid} score on \cifar w.r.t. different choices of noise schedule.}
    \vspace{1ex}

    \centering

    \resizebox{\textwidth}{!}{
        \setlength{\tabcolsep}{2pt}
        \setlength\extrarowheight{2.2pt}
        \begin{sc}
            \small
            \rowcolors{2}{}{mylightgray}

            \begin{tabular}{r||c|ccc}
                \toprule
                Model       & \multicolumn{1}{c|}{vanilla} & \multicolumn{1}{c}{sigmoid} & \multicolumn{1}{c}{cosine} & \multicolumn{1}{c}{linear} \\
                \midrule
                \midrule
                real-nvp    & $191.98$                     & $121.75$                    & $118.71$                   & $123.93$                   \\
                glow        & $74.62$                      & $64.87$                     & $71.90$                    & $74.36$                    \\
                \midrule
                vae         & $191.98$                     & $155.13$                    & $154.71$                   & $156.47$                   \\
                $\beta$-vae & $112.42$                     & $93.90$                     & $92.86$                    & $93.14$                    \\
                iwae        & $183.04$                     & $146.70$                     & $146.49$                    & $149.16$                    \\
                \bottomrule
            \end{tabular}
        \end{sc}
    }
    \label{tab:noise_schedule}
\end{table}

\end{minipage}

\paragraph{Comparisons with the two-step approach in \cite{LoaizaGanem2022diagnosing}.}
Manifold overfitting in likelihood-based \glspl{GM} has been recently analyzed in \cite{LoaizaGanem2022diagnosing}, which provides a two-step procedure to mitigate the issue.
The first step maps inputs into a low-dimensional space to handle the intrinsic low dimensionality of the data.
This step is then followed by likelihood-based density estimation on the resulting lower-dimensional representation.
This is achieved by means of a generalized autoencoder, which relies on a certain set of explicit deep \glspl{GM}, such as \glspl{VAE}.
Here, we compare our proposal with this two-step approach; results are reported in \cref{tab:two_step}.
To ensure a fair comparison, we use the same network architecture for our \gls{VAE} and their generalized autoencoder, and we rely on their official implementation \footnote{\url{https://github.com/layer6ai-labs/two_step_zoo}}.
Following \cite{LoaizaGanem2022diagnosing}, we consider a variety of density estimators in the low-dimensional space such as \glspl{NF}, \glspl{ARM} \citep{Uria2013}, \gls{AVB} \citep{Mescheder2017} and \glspl{EBM} \citep{Du2019}.
We observe that Gaussian mollification is better or comparable with these variants. 
In addition, our method is extremely simple and readily applicable to any likelihood-based \glspl{GM} without any extra auxilary models or the need to modify training procedures.

\begin{table}[H]
    \caption{Comparisons of \textsc{fid} scores on \cifar between mollitication and two-step methods.}
    \vspace{0.5ex}

    \centering

    \resizebox{0.9\columnwidth}{!}{
        \setlength{\tabcolsep}{5pt}
        \setlength\extrarowheight{1.5pt}
        \begin{sc}
            \small
            \rowcolors{3}{}{mylightgray}

            \begin{tabular}{c||cc|cccc}
                \toprule
                \multirow{2}{*}{vanilla vae} & \multicolumn{2}{c|}{mollification} & \multicolumn{4}{c}{two-step}                                           \\ \cline{2-7}
                                             & vae+gaussian                       & vae + blurring               & vae+nf  & vae+ebm  & vae+avb  & vae+arm \\
                \midrule
                \midrule
                191.98                       & $155.13$                             & $175.40$                        & $208.80$ & $166.20$ & $153.72$ & $203.32$ \\
                \bottomrule
            \end{tabular}
        \end{sc}
    }
    \label{tab:two_step}
\end{table}

\subsection{Density Estimation on UCI Data Sets} \label{sec:density_estimation}

We further evaluate our proposed method  in the context of density estimation tasks using \textsc{uci} data sets \citep{asuncion2007uci}, following \citep{Song20a}.
We consider three different types of normalizing flows: masked autorgressive flows (\textsc{maf}) \citep{Papamakarios2017}, \textsc{real-nvp} \citep{Dinh2017density} and \textsc{glow} \citep{Kingma2018Glow}.
To ensure a fair comparison, we apply the same experimental configurations for both the vanilla and the proposed method, including random seeds, network architectures, optimizer, and the total number of iterations.
\cref{tab:fid} shows the average log-likelihood (the higher the better) on the test data.
Error bars correspond to the standard deviation computed over 4 runs.
As it can be seen, our proposed Gaussian mollification approach consistently and significantly outperforms vanilla training across all models and all datasets. 

\begin{table}[H]
    \caption{The average test log-likelihood (\emph{higher is better}) on the \textsc{uci} data sets.
    Error bars correspond to the standard deviation over 4 runs.}
    \vspace{1ex}

    \centering

    \resizebox{\columnwidth}{!}{
        \setlength{\tabcolsep}{3.5pt}
        \setlength\extrarowheight{2.2pt}
        \begin{sc}
            \small
            \rowcolors{3}{}{mylightgray}

            \begin{tabular}{r||cc|cc|cc}
                \toprule
                \multirow{2}{*}{Dataset} & \multicolumn{2}{c|}{maf \citep{Papamakarios2017}}        & \multicolumn{2}{c|}{real-nvp \citep{Dinh2017density}}   & \multicolumn{2}{c}{glow \citep{Kingma2018Glow}}                                                                                                          \\ \cline{2-7}
                                         & vanilla                        & mollification                  & vanilla                        & mollification                  & vanilla                        & mollification                  \\
                \midrule
                \midrule
                red-wine                  & -16.32 {\scriptsize$\pm$ 1.88} & -11.51 {\scriptsize$\pm$ 0.44} & -27.83 {\scriptsize$\pm$ 2.56} & -12.51 {\scriptsize$\pm$ 0.40} & -18.21 {\scriptsize$\pm$ 1.14} & -12.37 {\scriptsize$\pm$ 0.33} \\
                white-wine                & -14.87 {\scriptsize$\pm$ 0.24} & -11.96 {\scriptsize$\pm$ 0.17} & -18.34 {\scriptsize$\pm$ 2.77} & -12.30 {\scriptsize$\pm$ 0.16} & -15.24 {\scriptsize$\pm$ 0.69} & -12.44 {\scriptsize$\pm$ 0.36} \\
                parkinsons               & -8.27  {\scriptsize$\pm$ 0.24} & -6.17  {\scriptsize$\pm$ 0.17} & -14.21 {\scriptsize$\pm$ 0.97} & -7.74  {\scriptsize$\pm$ 0.27} & -8.29  {\scriptsize$\pm$ 1.18} & -6.90  {\scriptsize$\pm$ 0.24} \\
                miniboone                & -13.03 {\scriptsize$\pm$ 0.04} & -11.65 {\scriptsize$\pm$ 0.09} & -20.01 {\scriptsize$\pm$ 0.22} & -13.96 {\scriptsize$\pm$ 0.12} & -14.48 {\scriptsize$\pm$ 0.10} & -13.88 {\scriptsize$\pm$ 0.08} \\
                \bottomrule
            \end{tabular}
        \end{sc}
    }
    \label{tab:uci}
\end{table}

\paragraph{Limitations.}

One limitation is that this work focuses exclusively on likelihood-based \glspl{GM}.
On image data, the improvements in \gls{FID} score indicate that the performance boost is generally substantial, but still far from being comparable with state-of-the-art \glspl{DM}.
While this may give an impression of a low impact, we believe that this work is important in pointing to one of the successful aspects characterizing \glspl{DM} and show how this can be easily integrated in the optimization of likelihood-based \glspl{GM}.
A second limitation is that, in line with the literature on \glspl{GM} for image data, where models are extremely costly to train and evaluate, we did not provide error bars on the results reported in the tables in the experimental section.
Having said that, the improvements reported in the experiments have been shown on a variety of models and on two popular image data sets.
Furthermore, the results are supported theretically and experimentally by a large literature on continuation methods for optimization.

\paragraph{Broader impact.}
This work provides an efficient way to improve a class of \glspl{GM}.
While we focused mostly on images, the proposed method can be applied to other types of data as shown in the density estimation experiments on the \textsc{uci} datasets.
Like other works in this literature, the proposed method can have both positive (e.g., synthesizing new data automatically or anomaly detection) and negative (e.g., deep fakes) impacts on society depending on the application.

\section{Related work} \label{sec:related}
Our work is positioned within the context of improving \glspl{GM} through the introduction of noise to the data.
One popular approach is the use of denoising autoencoders \citep{Vincent2008}, which are trained to reconstruct clean data from noisy samples. 
Building upon this, \cite{Bengio14} proposed a framework for modeling a Markov chain whose stationary distribution approximates the data distribution. 
In addition, \cite{Vincent2011} showed a connection between denoising autoencoders and score matching, which is an objective closely related to recent diffusion models \citep{SohlDickstein15, Ho2020}.
More recently, \cite{Meng2021} introduced a two-step approach to improve autoregressive generative models, where a smoothed version of the data is first modeled by adding a fixed level of noise, and then the original data distribution is recovered through an autoregressive denoising model.
In a similar vein, \cite{LoaizaGanemNeurIPSworkshop2022} recently attempted to use Tweedie's formula \citep{Robbins1956empirical} as a denosing step, but surprisingly found that it does not improve the performance of \glspl{NF} and \glspl{VAE}. 
Our work is distinct from these approaches in that Gaussian mollification guides the estimated distribution towards the true data distribution in a progressive manner by means of annealing instead of fixing a noise level.
Moreover, our approach does not require any explicit denoising step, and it can be applied off-the-shelf to the optimization of any likelihood-based \glspl{GM} without any modifications.

\section{Conclusion} \label{sec:conclusions}

Inspired by the enormous success of score-based \glsreset{DM} \glspl{DM}, in this work we hypothesized that data mollification is partially responsible for their impressive performance in generative modeling tasks.
In order to test this hypothesis, we introduced data mollification within the optimization of likelihood-based \glsreset{GM} \glspl{GM}, focusing in particular on \glsreset{NF} \glspl{NF} and \glsreset{VAE} \glspl{VAE}.
Data mollification is extremely easy to implement and it has nice theoretical properties due to its connection with continuation methods in the optimization literature, which are well-known techniques to improve optimization.
We applied this idea to challenging generative modeling tasks involving imaging data and relatively large-scale architectures as a means to demonstrate systematic gains in performance in various conditions and input dimensions.
We measured performance in quality of generated images through the popular \gls{FID} score.

While we are far from closing the gap with \glspl{DM} in achieving competing \gls{FID} score, we are confident that this work will serve as the basis for future works on performance improvements in state-of-the-art models mixing \glspl{DM} and likelihood-based \glspl{GM}, and in alternative forms of mollification to improve optimization of state-of-the-art \glspl{GM}.
For example, it would be interesting to study how to apply data mollification to improve the training of \glspl{GAN}; preliminary investigations show that the strategy proposed here does not offer significant performance improvements, and we believe this is due to the fact that data mollification does not help in smoothing the adversarial objective.
Also, while our study shows that the data mollification schedule is not critical, it would be interesting to study whether it is possible to derive optimal mollification schedules, taking inspiration, e.g., from \citep{Iwakiri2022}.
We believe it would also be interesting to consider mixture of likelihood-based \glspl{GM} to counter problems due to the union of manifolds hypothesis, whereby the intrinsic dimension changes over the input \citep{BrownICLR2023}.
Finally, it would be interesting to investigate other data, such as 3D point cloud data \citep{Yang2019} and extend this work to deal with other tasks, such as supervised learning.

\begin{ack}
	MF gratefully acknowledges support from the AXA Research Fund and the Agence Nationale de la Recherche (grant ANR-18-CE46-0002 and ANR-19-P3IA-0002).
\end{ack}

{
\bibliographystyle{abbrv}
\bibliography{bib/main_biblio.bib}
}

\newpage
\appendix
\section{A Primer on Normalizing Flows and VAEs}

Given a dataset $\mathcal{D}$ consisting of $N$ \iid samples $\cD \stackrel{\Delta}{=} \{ \mbx_{i} \}_{i=1}^{N}$ with $\mbx_{i} \in \mathbb{R}^{D}$, we aim at estimating the unknown continuous generating distribution $p_{\text{data}}(\mbx)$.
In order to do so, we introduce a model $p_{\mbtheta}(\mbx)$ with parameters $\mbtheta$ and attempt to estimate $\mbtheta$ based on the dataset $\mathcal{D}$.
A common approach to estimate $\mbtheta$ is to maximize the likelihood  of the data, which is equivalent to minimizing the following objective:
\begin{align}
    \cL(\mbtheta) &\stackrel{\Delta}{=}   - \mathbb{E}_{p_{\text{data}}(\mbx)} \left[ \log p_{\mbtheta}(\mbx) \right] \label{eq:obj_likelihood}. %
\end{align}
Optimization for this objective can be done through a stochastic gradient descent algorithm using minibatches of samples from $p_{\text{data}}(\mbx)$.

\subsection{Normalizing Flows}
In flow-based generative models \citep{Papamakarios2021, Kobyzev2021}, the generative process is defined as:
\begin{align}
    \mbz \sim p_{\mbphi}(\mbz); \quad \mbx = \mbf_{\mbpsi}(\mbz),
\end{align}
where $\mbz \in \mathbb{R}^{D}$ is a latent variable, and $p_{\mbphi}(\mbz)$ is a tractable base distribution with parameters $\mbphi$, such as an isotropic multivariate Gaussian.
The function $\mbf_{\mbpsi}: \mathbb{R}^{D} \rightarrow \mathbb{R}^{D} $ is invertible, such that given any input vector $\mbx$ we have $\mbz = \mbf^{-1}_{\mbpsi}(\mbx)$.
A \glsreset{NF}\gls{NF} \citep{Rezende2015} defines a sequence of invertible transformations $\mbf = \mbf_1 \circ \mbf_2 \circ \cdots \mbf_K$, such that the relationship between $\mbx$ and $\mbz$ can be written as:
\begin{equation}
    \mbx  \stackrel{\mbf_1}{\longleftrightarrow} \mbh_1 \stackrel{\mbf_2}{\longleftrightarrow} \mbh_2 \cdots  \stackrel{\mbf_K}{\longleftrightarrow} \mbz,
\end{equation} 
where $\mbh_k = \mbf_k^{-1}(\mbh_{k-1}; \mbpsi_{k})$ and $\mbpsi_{k}$ are the parameters of the transformation $\mbf_k$.
For the sake of simplicity, we define $\mbh_0 \stackrel{\Delta}{=} \mbx$ and $\mbh_K \stackrel{\Delta}{=} \mbz$.
The likelihood  of the model given a datapoint can be computed analytically using the change of variables as follows:
\begin{align}
    \log p_{\mbtheta} (\mbx) &= \log p_{\mbphi}(\mbz) + \log \left|  \det (\partial\mbz / \partial\mbx) \right| \\
    &= \log p_{\mbphi}(\mbz) + \sum_{k=1}^{K} \log \left|  \det (\partial\mbh_k  / \partial\mbh_{k-1}) \right|,
\end{align}
where $\log | \det (\partial\mbh_k / \partial \mbh_{k-1}) |$ is the logarithm of absolute value of the determinant of the Jacobian matrix $\partial\mbh_k / \partial \mbh_{k-1}$.
This term accounts for the change of measure when going from $\mbh_{k-1}$ to $\mbh_{k}$ using the transformation $\mbf_k$.
The resulting \gls{NF} model is then characterized by the set of parameters $\mbtheta = \{ \mbphi \} \cup \{ \mbpsi_{k} \}_{k=1}^{K}$, which can be estimated using the \gls{MLE} objective \cref{eq:obj_likelihood}. %

Though \glspl{NF} allow for exact likelihood computation, they require $\mbf_{k}$ to be invertible and to have a tractable inverse and Jacobian determinant.
This restricts the flexibility to certain transformations that can be used within \glspl{NF} \cite[see e.g.,][and references therein]{Papamakarios2021,Kobyzev2021}, such as affine coupling \citep{Dinh2014nice,Dinh2017density}, invertible convolution \citep{Kingma2018Glow}, spline \citep{Durkan2019NeuralSF,Dolatabadi20a}, or inverse autoregressive transformations \citep{Kingma2016}.

\subsection{Variational Autoencoders}
\glsreset{VAE} \glspl{VAE} \citep{KingmaW13,Rezende2015} introduce a low-dimensional latent variable $\mbz \in \mathbb{R}^{P}$, with $P \ll D$, to the generative process as follows:
\begin{align}
    \mbz \sim p(\mbz); \quad \mbx \sim p_{\mbtheta}(\mbx \g \mbz).
\end{align}
Here, $p(\mbz)$ is a tractable prior distribution over the latent variables $\mbz$, and $p_{\mbtheta}(\mbx \g \mbz)$, which is also known as a decoder, is usually implemented by a flexible neural network parameterized by $\mbtheta$.
Different from \glspl{NF}, \glspl{VAE} employ a stochastic transformation $p_{\mbtheta}(\mbx \g \mbz)$ to map $\mbz$ to $\mbx$. %
Indeed, \glspl{NF} can be viewed as \glspl{VAE} where the decoder and encoder are modelled by Dirac deltas $p_{\mbtheta}(\mbx \g \mbz) = \delta\big(\mbf_{\mbtheta}(\mbx)\big)$ and $q_{\mbphi}(\mbz \g \mbx) = \delta\big(\mbf^{-1}_{\mbtheta}(\mbx)\big)$ respectively, using a restricted family of transformations $\mbf_{\mbtheta}$.

The marginal likelihood of \glspl{VAE} is intractable and given by:
\begin{equation}
    p_{\mbtheta}(\mbx) = \int p_{\mbtheta}(\mbx \g \mbz) p(\mbz) d\mbz.
\end{equation}
A variational lower bound on ther marginal likelihood can be obtained by introducing a variational distribution $q_{\mbphi}(\mbz \g \mbx)$, with parameters $\mbphi$, which acts as an approximation to the the unknown posterior $p(\mbz \g \mbx)$:
\begin{equation}
    \log p_{\mbtheta}(\mbx) \geq \underbrace{\mathbb{E}_{q_{\mbphi}(\mbz \g \mbx)} [ \log p_{\mbtheta}(\mbx \g \mbz)] - \KL{q_{\mbphi}(\mbz \g \mbx)}{p(\mbz)}}_{\cL_{\text{ELBO}}(\mbtheta, \mbphi)},
\end{equation}
where, $\cL_{\text{ELBO}}(\mbtheta, \mbphi)$ is known as the \gls{ELBO}, and the expectation can be approximated by using Monte Carlo samples from $q_{\mbphi}(\mbz \g \mbx)$.
This objective can be optimized with stochastic optimization w.r.t. parameters $\mbtheta$ and $\mbphi$ in place of \cref{eq:obj_likelihood}.

To tighten the gap between the \gls{ELBO} and the true marginal likelihood, \glspl{VAE} can be employed with an expressive form of the approximate posterior $q_{\mbphi}(\mbz \g \mbx)$ such as importance weighted sampling \citep{Burda2015importance} or normalizing flows \citep{Kingma2016,BergHTW18}.
In addition, to avoid the over regularization offect induced by the prior $p(\mbz)$, one can utilize a flexible prior such as multi-modal distributions \citep{Dilokthanakul2016deep,Tomczak18a}, hierarchical forms \citep{Sonerby2016ladder, Klushyn2019}, or simply reweighing the KL divergence term in the \gls{ELBO} \citep{Higgins2017betavae}.
\section{Details on Blurring Mollification} \label{appendix:blur_mollification}

Recently, \cite{Rissanen2023generative,Hoogeboom2023blurring,Daras2022soft} have proposed appproaches to destroy information of images using blurring operations for diffusion-type generative models. 
Their approach involves stochastically reversing the heat equation, which is a \gls{PDE} that can be used to erase fine-scale information when applied locally to the 2D plane of an image.
In particular, the Laplace \gls{PDE} for heat diffusions is as follows:
\begin{equation}
    \frac{\partial}{\partial t}\tilde{\mbx}(i,j,t) = \Delta \tilde{\mbx}(i,j,t), \label{eq:blurring_pde}
\end{equation}
where we consider the initial state of the system to be $\mbx$, the true image data.
This \gls{PDE} can be effectively solved by employing a diagonal matrix within the frequency domain of the cosine transform, provided that the signal is discretized onto a grid.
The solution to this equation at time $t$ can be effectively computed by:
\begin{equation}
    \tilde{\mbx}_t = \mbA_t \mbx = \mbV \mbD_t \mbV^{\top} \mbx, \label{eq:blurring_mollification} 
\end{equation}
Here, $\mbV^{\top}$ and $\mbV$ denote the \gls{DCT} and inverse \gls{DCT}, respectively; the diagonal matrix $\mbD_t$ is the exponent of a weighting matrix for frequencies $\mbLambda$ so that $\mbD_t = \exp(\mbLambda t)$.
We refer the reader to Appendix A of \cite{Rissanen2023generative} for the specific definition of $\mbLambda$.
We can evaluate \cref{eq:blurring_mollification} in the Fourier domain, which is fast to compute, as the \gls{DCT} and inverse \gls{DCT} require $\mathcal{O}(N \log N)$ operations.
The equivalent form of \cref{eq:blurring_mollification} in the Fourier domain is as follows:
\begin{equation}
    \tilde{\mbu}_t = \exp(\mbLambda t) \mbu, \label{eq:blurring_mollification_fourier} 
\end{equation}
where $\mbu = \mbV^{\top} \mbx = \text{DCT}(\mbx)$.
As $\mbLambda$ is a diagonal matrix, the above Fourier-space model is fast to evaluate.
A Python implementation of this blurring mollification is presented in \cref{alg:python_blurring_mollification}.

We follow \cite{Rissanen2023generative} to set the schedule for the blurring mollification.
In particular, we use a logarithmic spacing for the time steps $t_k$, where $t_{0} = \nicefrac{\sigma^{2}_{B,\text{max}}}{2}$ and $t_{T} = \nicefrac{\sigma^{2}_{B,\text{min}}}{2} = \nicefrac{0.5^2}{2}$,  corresponding to sub-pixel-size blurring.
Here, $\sigma^{2}_{B,\text{max}}$ is the effective lengthscale-scale of blurring at the beginning of the mollification process.
Following \cite{Rissanen2023generative}, we set this to half the width of the image.

\definecolor{codegreen}{rgb}{0,0.6,0}
\definecolor{codegray}{rgb}{0.5,0.5,0.5}
\definecolor{codepurple}{rgb}{0.58,0,0.82}
\definecolor{backcolour}{rgb}{0.95,0.95,0.92}
\lstdefinestyle{mystyle}{
    backgroundcolor=\color{white},
    numberstyle=\tiny\color{codegray},
    stringstyle=\color{codepurple},
    basicstyle=\fontsize{7.5pt}{7.5pt}\ttfamily\selectfont,
    commentstyle=\fontsize{7.5pt}{7.5pt}\color{codegreen},
    keywordstyle=\fontsize{7.5pt}{7.5pt}\color{magenta},
    breakatwhitespace=false,
    breaklines=true,
    captionpos=b,
    keepspaces=true,
    numbers=left,
    numbersep=5pt,
    showspaces=false,
    showstringspaces=false,
    showtabs=false,
    tabsize=2
}
\lstset{style=mystyle}

\begin{algorithm}[h]
    \centering
    \caption{Python code for blurring mollification}\label{alg:python_blurring_mollification}
    \begin{lstlisting}[language=Python]
import numpy as np
from scipy.fftpack import dct, idct

def blurring_mollify(x, t):
    # Assuming the image u is an (KxK) numpy array
    K = x.shape[-1]
    freqs = np.pi*np.linspace(0,K-1,K)/K
    frequencies_squared = freqs[:,None]**2 + freqs[None,:]**2
    x_proj = dct(u, axis=0, norm='ortho')
    x_proj = dct(x_proj, axis=1, norm='ortho')
    x_proj = np.exp(-frequencies_squared * t) * x_proj
    x_mollified = idct(x_proj, axis=0, norm='ortho')
    x_mollified = idct(x_mollified, axis=1, norm='ortho')
    return x_mollified
\end{lstlisting}%
\end{algorithm}

\section{Implementation of Noise Schedules} \label{appendix:noise_schedules}

\cref{alg:python_noise_schedules} shows the Python code for the noise schedules used in this work.
For the sigmoid schedule, following \cite{Chen2023}, we set the default values of {\small \texttt{start}} and {\small \texttt{end}} to $0$ and $3$, respectively. %

\definecolor{codegreen}{rgb}{0,0.6,0}
\definecolor{codegray}{rgb}{0.5,0.5,0.5}
\definecolor{codepurple}{rgb}{0.58,0,0.82}
\definecolor{backcolour}{rgb}{0.95,0.95,0.92}
\lstdefinestyle{mystyle}{
    backgroundcolor=\color{white},
    numberstyle=\tiny\color{codegray},
    stringstyle=\color{codepurple},
    basicstyle=\fontsize{7.5pt}{7.5pt}\ttfamily\selectfont,
    commentstyle=\fontsize{7.5pt}{7.5pt}\color{codegreen},
    keywordstyle=\fontsize{7.5pt}{7.5pt}\color{magenta},
    breakatwhitespace=false,
    breaklines=true,
    captionpos=b,
    keepspaces=true,
    numbers=left,
    numbersep=5pt,
    showspaces=false,
    showstringspaces=false,
    showtabs=false,
    tabsize=2
}
\lstset{style=mystyle}

\begin{algorithm}[h]
    \centering
    \caption{Python code for noise schedules}\label{alg:python_noise_schedules}
    \begin{lstlisting}[language=Python]
import numpy as np

def sigmoid(x):
    # Sigmoid function.
    return 1 / (1 + np.exp(-x))

def sigmoid_schedule(t, T, tau=0.7, start=0, end=3, clip_min=1e-9):
    # A scheduling function based on sigmoid function with a temperature tau.
    v_start = sigmoid(start / tau)
    v_end = sigmoid(end / tau)
    return (v_end - sigmoid((t/T * (end - start) + start) / tau)) / (v_end - v_start)

def linear_schedule(t, T):
    # A scheduling function based on linear function.
    return 1 - t/T

def cosine_schedule(t, T, ns=0.0002, ds=0.00025):
    # A scheduling function based on cosine function.
    return np.cos(((t/T + ns) / (1 + ds)) * np.pi / 2)**2


\end{lstlisting}%
\end{algorithm}

\section{Experimental Details} \label{appendix:experiment_details}

\subsection{Data sets}

\paragraph{Synthetic data sets.}
\begin{itemize}
    \item \emph{Mixture of Gaussians}: We consider a mixture of two Gaussians with means $\mu_{k} = (2 \sin(\pi k) , 2 \cos(\pi k) )$ and covariance matrices $\mbSigma_{k} = \sigma^{2} \mbI$, where $\sigma = \frac{2}{3} \sin(\pi / 2)$.
    We generate 10K samples for training and 10K samples for testing from this distribution.
    \item \emph{Von Mises distribution:} We use a von Mises distribution with parameters $\kappa = 1$, and then transform to Cartesian coordinates to obtain a distribution on the unit circle in $\mathbb{R}^{2}$. 
    We generate 10K training samples and 10K testing from this distribution.

\end{itemize}

\paragraph{Image data sets.}
We consider two image data sets including \cifar \citep{Krizhevsky2009learning} and \celeba \citep{LiuLWT15}.
These data sets are publicly available and widely used in the literature of generative models.
We use the official train/val/test splits for both data sets.
The resolution of \cifar is $3 \times 32 \times 32$.
For \celeba, we pre-process images by first taking a $148 \times 148$ center crop and then resizing to the $3 \times 64 \times 64$ resolution.

\paragraph{UCI data sets.}

We consider four data sets in the \textsc{uci} repository \citep{asuncion2007uci}: \textsc{red-wine}, \textsc{white-wine}, \textsc{parkinsons}, and \textsc{miniboone}.
$10$\% of the data is set aside as a test set, and an additional $10$\% of the remaining data is used for validation.
To standardize the features, we subtract the sample mean from each data point and divide by the sample standard deviation.

\subsection{Software and Computational Resources}
We use NVIDIA P100 and A100 GPUs for the experiments, with 16GB and 80GB of memory respectively. 
All models are trained on a single GPU except for the experiments with \textsc{nvae} model \citep{Vahdat2020}, where we employ two A100 GPUs.
We use PyTorch \citep{Paszke2019} for the implementation of the models and the experiments.
Our experiments with \glspl{VAE} and \glspl{NF} are relied on the \texttt{pythae} \citep{Chadebec2022} and \texttt{normflows} \citep{Vincent2023} libraries, respectively.

\subsection{Training Details}

\subsubsection{Toy examples.} \label{appendix:toy_details}
In the experiments on synthetic data sets, we use a \textsc{real-nvp} flow \citep{Dinh2017density} with $5$ affine coupling layers consisting of $2$ hidden layers of $64$ units each.
We train the model for $20000$ itereations using an Adam optimizer \citep{Kingma2015Adam} with a learning rate of $5\cdot10^{-4}$ and a mini-batch size of $256$.

\subsubsection{Imaging experiments.}

\paragraph{\textsc{real-nvp}.}
We use the multi-scale architecture with deep convolutional residual networks in the coupling layers as described in \cite{Dinh2017density}.
For the \cifar data set, we use $4$ residual blocks with $32$ hidden feature maps for the first coupling layers with checkerboard masking.
For the \celeba data set, $2$ resdiual blocks are employed.
We use an Adam optimizer \citep{Kingma2015Adam} with a learning rate of $10^{-3}$ and a mini-batch size of $64$.
We train the model for $100$ and  $80$ epochs on the \cifar and \celeba data sets, respectively.
For the mollification training, we perturb the data for $50$ and $40$ epochs for \cifar and \celeba, respectively.

\paragraph{\textsc{glow}.}
We use a multi-scale architecture as described in \cite{Kingma2018Glow}.
The architecture has a depth level of $K=20$, and a number of levels $L=3$.
We use the AdaMax \citep{Kingma2015Adam} optimizer with a learning rate of $3\cdot 10^{-4}$ and a mini-batch size of $64$.
We train the model for $80$ and $40$ epochs on the \cifar and \celeba data sets, respectively.
For the mollification training, we perturb the data for $50$ and $20$ epochs for \cifar and \celeba, respectively.

\paragraph{\glspl{VAE}.}

\begin{table}[H]
    \caption{Neural network architectures used for \glspl{VAE} in our experiments.
        Here, {\scriptsize $\textsc{Conv}_{(n,s,p)}$} and {\scriptsize $\textsc{ConvT}_{(n,s,p)}$} respectively denotes convolutional layer and transposed convolutional layers with $n$ filters, a stride of $s$ and a padding of $p$, whereas {\scriptsize $\textsc{FC}_{n}$} represents a fully-connected layer with $n$ units, and  {\scriptsize BN} denotes a batch-normalization layer.
    }
    \vspace{1ex}

    \centering

    \resizebox{0.8\columnwidth}{!}{
        \setlength{\tabcolsep}{5pt}
        \setlength\extrarowheight{1.5pt}
        \begin{sc}
            \small

            \begin{tabular}{c  p{0.4\linewidth} p{0.4\linewidth}}
                \toprule
                                                                                                 & Cifar10 & CelebA \\
                \midrule
                \midrule
                Encoder: 
                         & $\mbx \in \mathbb{R}^{3{\times}32{\times}32}$ \newline
                $\rightarrow \text{Conv}_{(128,4,2)} \rightarrow \text{BN} \rightarrow \text{ReLU}$\newline
                $\quad \rightarrow \text{Conv}_{(256,4,2)} \rightarrow \text{BN} \rightarrow \text{ReLU}$\newline
                $\quad \rightarrow \text{Conv}_{(512,4,2)} \rightarrow \text{BN} \rightarrow \text{ReLU}$\newline
                $\quad \rightarrow \text{Conv}_{(1024,4,2)} \rightarrow \text{BN} \rightarrow \text{ReLU}$\newline
                $\quad \rightarrow \text{Flatten} \rightarrow \text{FC}_{256{\times}2}$
                        & $\mbx \in \mathbb{R}^{3{\times}64{\times}64}$ \newline
                $\rightarrow \text{Conv}_{(128,4,2)} \rightarrow \text{BN} \rightarrow \text{ReLU}$\newline
                $\quad \rightarrow \text{Conv}_{(256,4,2)} \rightarrow \text{BN} \rightarrow \text{ReLU}$\newline
                $\quad \rightarrow \text{Conv}_{(512,4,2)} \rightarrow \text{BN} \rightarrow \text{ReLU}$\newline
                $\quad \rightarrow \text{Conv}_{(1024,4,2)} \rightarrow \text{BN} \rightarrow \text{ReLU}$\newline
                $\quad \rightarrow \text{Flatten} \rightarrow \text{FC}_{256{\times}2}$                                      \\
          
                \midrule
                Decoder: 
                         & $\mbz \in \mathbb{R}^{256} \rightarrow \text{FC}_{8{\times}8{\times}1024}$\newline
                $\rightarrow \text{ConvT}_{(512,4,2)}\rightarrow \text{BN} \rightarrow \text{ReLU}$\newline
                $\rightarrow \text{ConvT}_{(256,4,2)}\rightarrow \text{BN} \rightarrow \text{ReLU}$\newline
                $\rightarrow \text{ConvT}_{(3,4,1)}$
                & $\mbz \in \mathbb{R}^{256} \rightarrow \text{FC}_{8{\times}8{\times}1024}$\newline
                $\rightarrow \text{ConvT}_{(512,5,2)}\rightarrow \text{BN} \rightarrow \text{ReLU}$\newline
                $\rightarrow \text{ConvT}_{(256,5,2)}\rightarrow \text{BN} \rightarrow \text{ReLU}$\newline
                $\rightarrow \text{ConvT}_{(128,5,2)}\rightarrow \text{BN} \rightarrow \text{ReLU}$\newline
                $\rightarrow \text{ConvT}_{(3,4,1)}$                                                                                \\
                \bottomrule
              \end{tabular}
        \end{sc}
    }
    \label{tab:vae_networks}
\end{table}
We use convolutional networks for both the encoder and decoder of \glspl{VAE} \citep{KingmaW13,RezendeMW14}.
\cref{tab:vae_networks} shows the details of the network architectures.
We use an Adam optimizer \citep{Kingma2015Adam} with a learning rate of $3 \cdot 10^{-4}$ and a mini-batch size of $128$.
We train the model for $200$ and $100$ epochs on the \cifar and \celeba data sets, respectively.
For the mollification training, we perturb the data for $100$ and $50$ epochs for \cifar and \celeba, respectively.
The addtional details of the variants of \glspl{VAE} are as follows:
\begin{itemize}
    \item \textsc{vae-iaf} \citep{Kingma2016}: We use a $3$-layer \textsc{made} \citep{Germain2015} with $128$ hidden units and \textsc{relu} activation for each layer and stack $2$ blocks of Masked Autoregressive Flow to create the flow for approximating the posterior.
    \item \textsc{$\beta$-vae} \citep{Higgins2017betavae}: We use a coefficient of $\beta = 0.1$ for the \gls{KL} term in the \gls{ELBO} objective.
    \item \textsc{iwae} \citep{Burda2015importance}: We use a number of importance samples of $K=5$.
    \item \textsc{hvae} \citep{Caterini2018}: We set the number of leapfrog steps to used in the integrator to $1$.
    The leapfrog step size is adaptive with an initial value of $0.001$
\end{itemize}

\paragraph{\textsc{nvae}.} 
We use the default network architecture as described in \cite{Vahdat2020}.
We train the model on the \cifar for $300$ epochs with an AdaMax optimizer \citep{Kingma2015Adam} with a learning rate of $10^{-3}$ and a mini-batch size of $200$.
For the mollification training, we perturb the data for first $150$ epochs.

\subsection{UCI experiments}

For \textsc{maf} models \citep{Papamakarios2017}, we employ $5$ autoregressive layers, each composed of a feedforward neural network utilizing masked weight matrices \citep{Germain2015}.
This neural networks consists of $512$ hidden units and employ the tanh activation function.
For \textsc{real-nvp} \citep{Dinh2017density} and \textsc{glow} \citep{Kingma2018Glow} models, we implement $5$ coupling layers, each comprising two feedforward neural networks with $512$ hidden units to handle the scaling and shift functions.
The first neural network utilizes the tanh activation function, while the latter employs a rectified linear activation function.
We introduce batch normalization \citep{Ioffe15} after each coupling layer in \textsc{real-nvp} and after each autoregressive layer in \textsc{maf}.
All models are trained with the Adam optimizer \cite{Kingma2015Adam} for $150$ epochs with a learning rate of $10^{-4}$ and a mini-batch size of $100$.
For mollification training, we employ Gaussian mollification during the entire training process.

\subsection{Evaluation Metrics}

\paragraph{Maximum Mean Discrepancy.}
The \gls{MMD} between two distributions $p_{\text{data}}$ and $p_{\mbtheta}$ is defined as follows \cite{Gretton12}:
\begin{equation}
    \text{MMD}(p_{\text{data}}, p_{\mbtheta}) = \sup_{\|h\|_{\mathcal{H}} \leq 1} \Big[ \mathbb{E}_{p_{\text{data}}}[h] - \mathbb{E}_{p_{\mbtheta}}[h] \Big],
\end{equation}
where $\mathcal{H}$ denotes a \gls{RKHS} induced by a characteristic kernel $K$.
The \gls{MMD} has the closed form:
\begin{equation}
    \text{MMD}^{2}(p_{\text{data}}, p_{\mbtheta}) = \mathbb{E}_{\mbx, \mbx' \sim p_{\text{data}}} [K(\mbx, \mbx')] + \mathbb{E}_{\mbx, \mbx' \sim p_{\mbtheta}} [K(\mbx, \mbx')] - 2 \mathbb{E}_{\mbx \sim p_{\text{data}}, \mbx' \sim p_{\mbtheta}} [K(\mbx, \mbx')],
\end{equation}
which can be estimated by using samples from $p_{\text{data}}$ and $p_{\mbtheta}$.
In our experiments, we use 10K test samples from the true data distribution $p_{\text{data}}$ and 10K samples from the model distribution $p_{\mbtheta}$ for estimating the \gls{MMD} score.

We employ the popular \gls{RBF} kernel for the \gls{MMD}, which is defined as follows:
\begin{equation}
    K(\mbx, \mbx') = \sigma^2 \exp \Big( - \frac{\| \mbx - \mbx' \|^2}{2 l} \Big),
\end{equation}
with a lengthscale $l=1$ and variance $\sigma^2 = 10^{-4}$.

\paragraph{FID score.}
To assess the quality of the generated images, we employed the widely used Fr\'echet Inception Distance \cite{HeuselRUNH17}.
The FID measures the Fr\'echet distance between two multivariate Gaussian distributions, one representing the generated samples and the other representing the real data samples. 
By comparing their distribution statistics, we can assess the similarity between the generated and real data distributions.
The FID score is defined as follows:
\begin{align}
    \text{FID} = \| \mu_{\text{real}} - \mu_{\text{gen}} \|^2 + \text{Tr}(\Sigma_{\text{real}} + \Sigma_{\text{gen}} - 2 \sqrt{\Sigma_{\text{real}} \Sigma_{\text{gen}}}).
\end{align}
The distribution statistics are obtained from the $2048$-dimensional activations of the pool3 layer of an Inception-v3 network.
We use the \texttt{pytorch-fid} \footnote{\url{https://github.com/mseitzer/pytorch-fid}} library for calculating the FID score in our experiments.

\section{Addtional Results} \label{appendix:additional_results}
\cref{tab:cifar10_figures} and \cref{tab:celeba_figures} illustrate uncurated samples from the trained models.
\cref{fig:cifar_fid_app} and \cref{fig:celeba_fid_app} show the progression of FID scores during training on the \cifar and \celeba datasets, respectively.

\newpage
\begin{table}[H]
    \caption{Uncurated samples from the models trained on the \cifar dataset.}
    \vspace{1ex}

    \centering

    \resizebox{\columnwidth}{!}{
        \setlength{\tabcolsep}{2pt}
        \begin{sc}
            \small

            \begin{tabular}{c  p{0.27\linewidth} p{0.27\linewidth} p{0.27\linewidth}}
                \toprule
                 & Vanilla & Gauss. Mollification & Blur. Mollification \\
                \midrule
                \midrule
                \raisebox{26pt}{real-nvp}
                 & \includegraphics[clip,width=\linewidth]{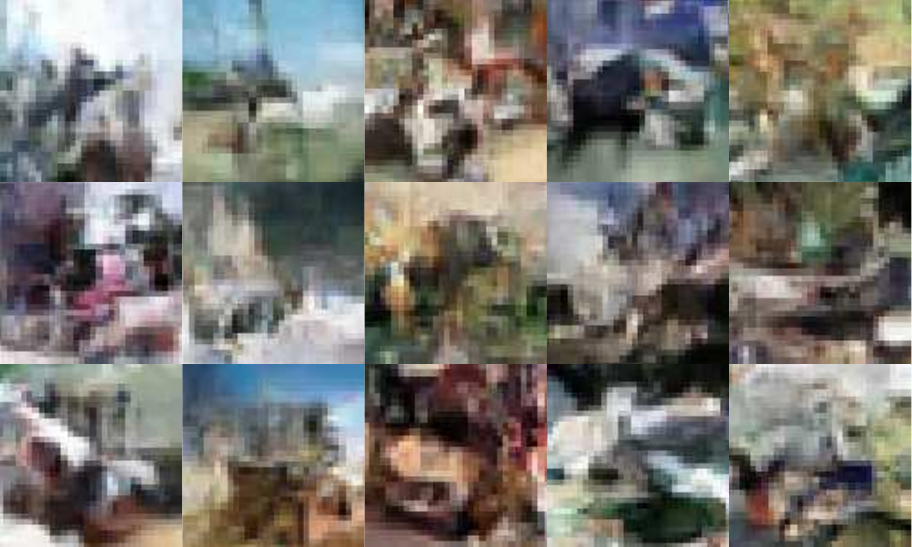}
                 & \includegraphics[clip,width=\linewidth]{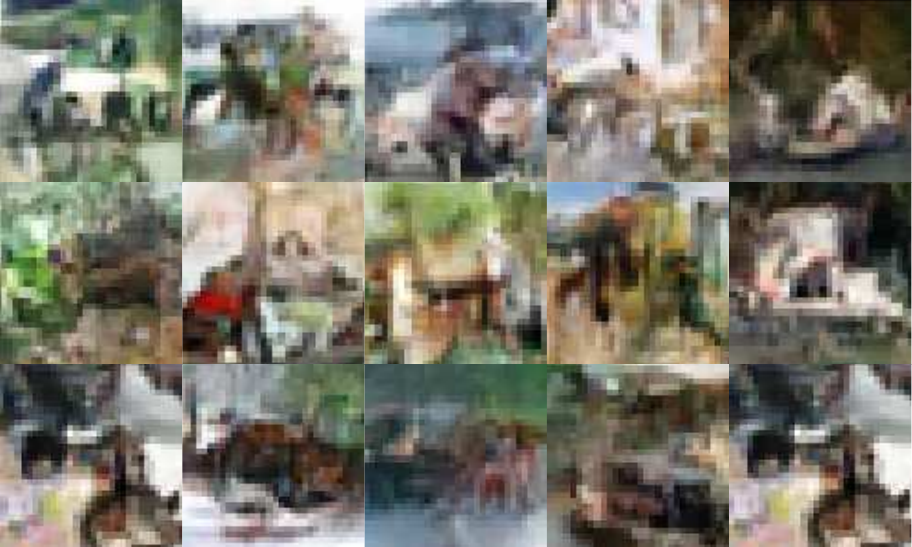}                 
                 & \includegraphics[clip,width=\linewidth]{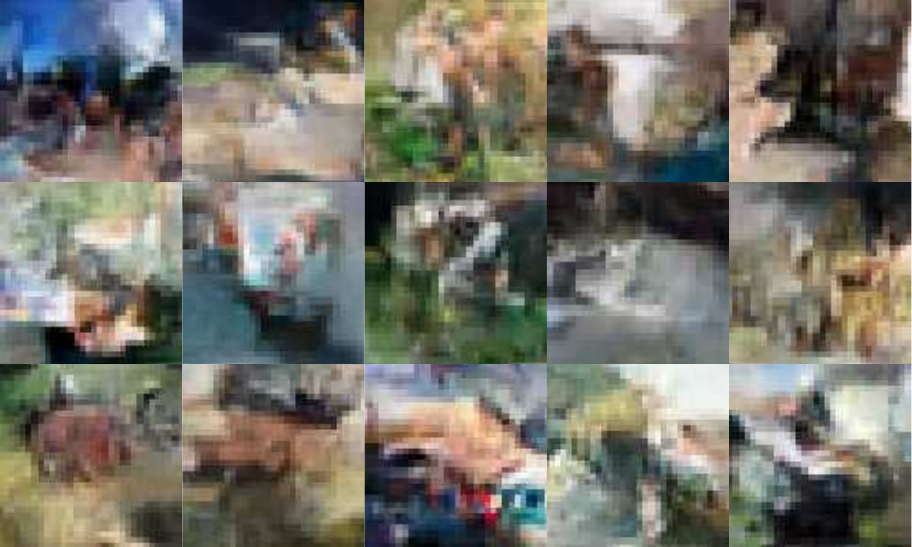}  \\

                \midrule
                \raisebox{26pt}{glow}
                 & \includegraphics[clip,width=\linewidth]{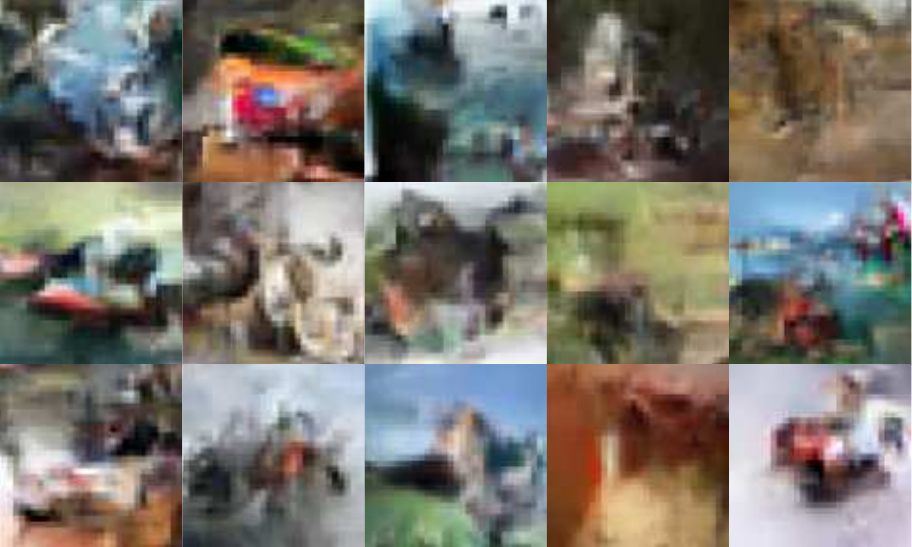}
                 & \includegraphics[clip,width=\linewidth]{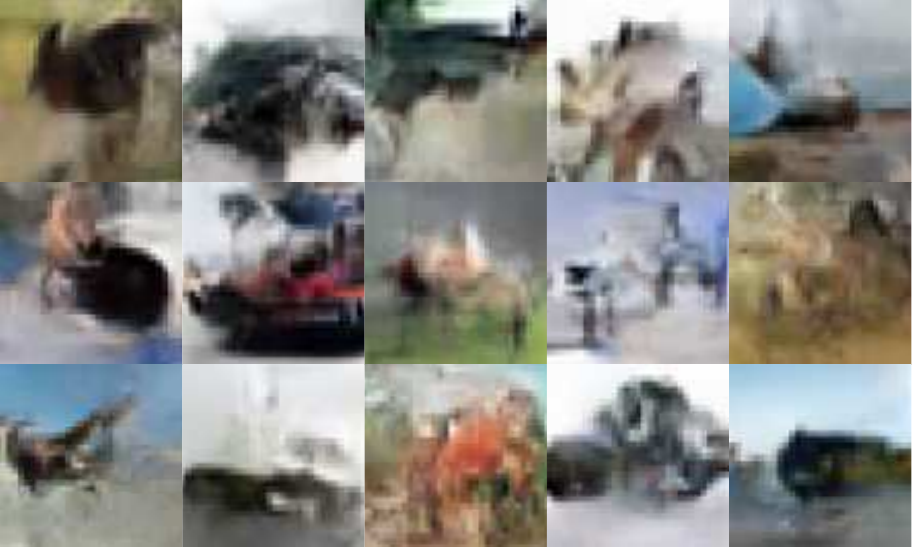}                 
                 & \includegraphics[clip,width=\linewidth]{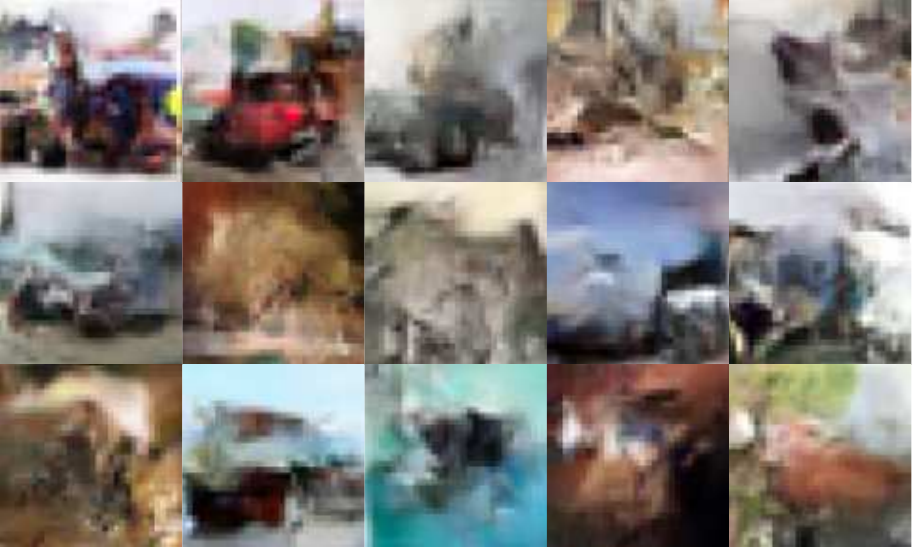}  \\

                 \midrule
                \raisebox{26pt}{vae}
                 & \includegraphics[clip,width=\linewidth]{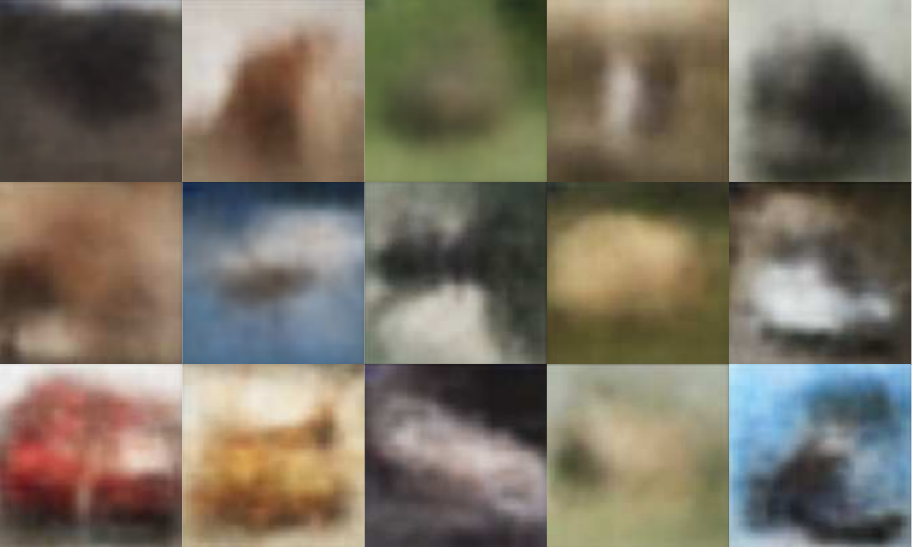}
                 & \includegraphics[clip,width=\linewidth]{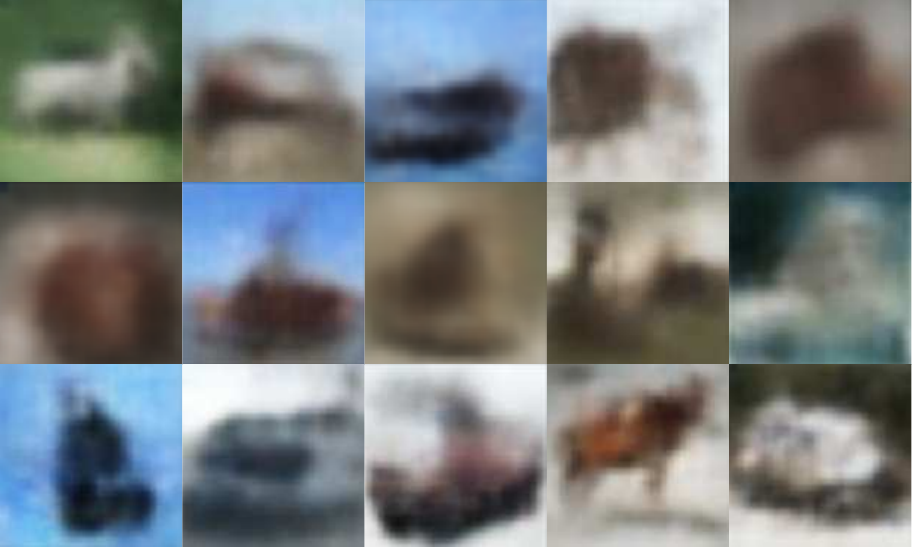}                 
                 & \includegraphics[clip,width=\linewidth]{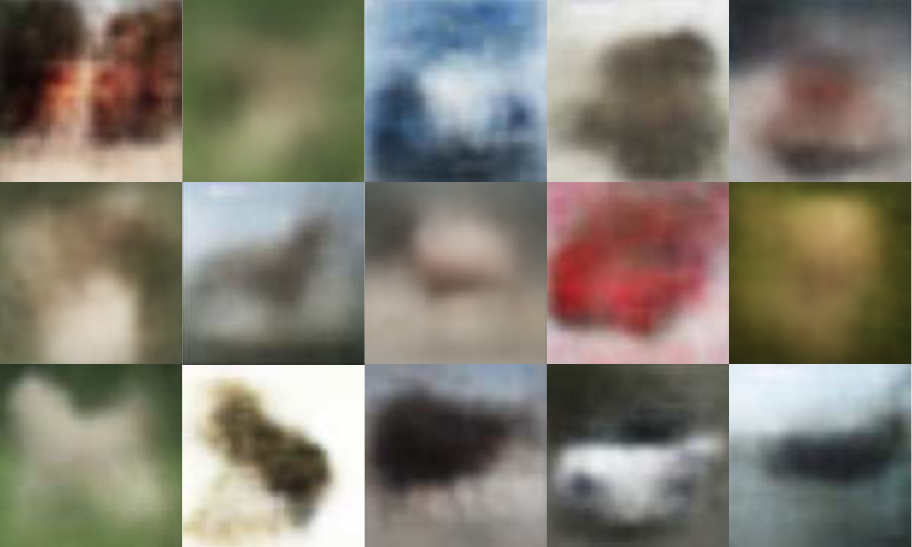}  \\

                 \midrule
                \raisebox{26pt}{vae-iaf}
                 & \includegraphics[clip,width=\linewidth]{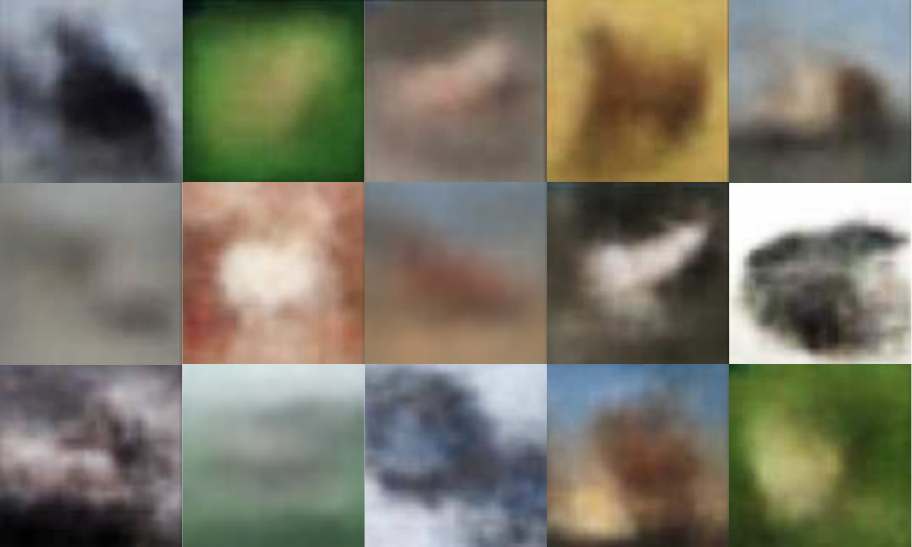}
                 & \includegraphics[clip,width=\linewidth]{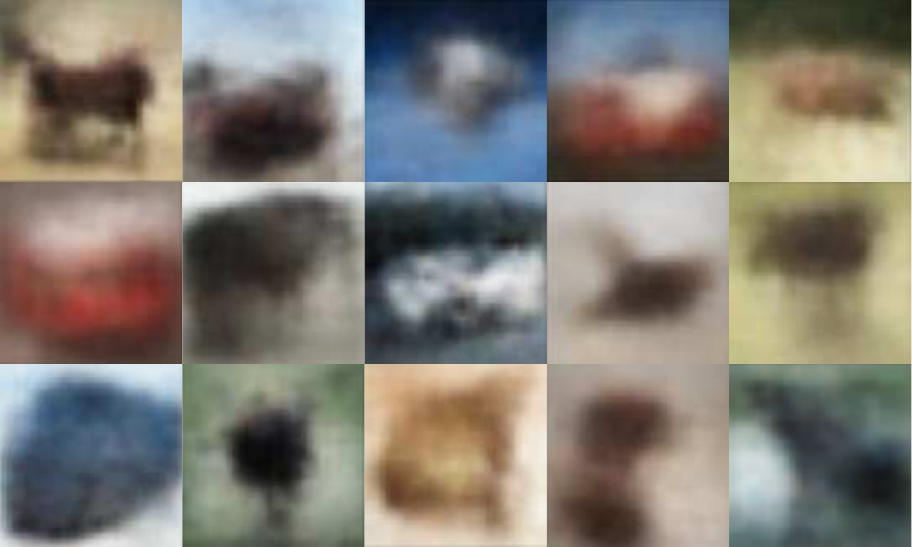}                 
                 & \includegraphics[clip,width=\linewidth]{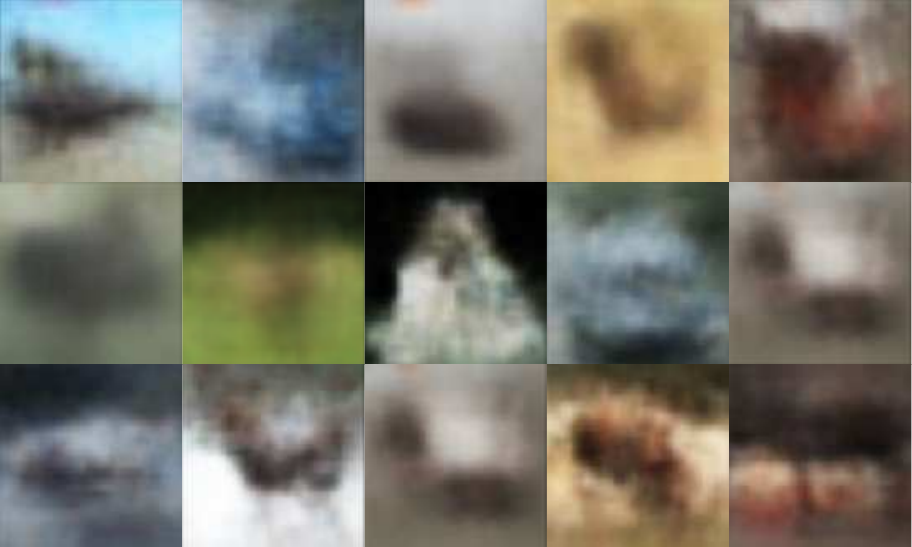}  \\

                 \midrule
                \raisebox{26pt}{iwae}
                 & \includegraphics[clip,width=\linewidth]{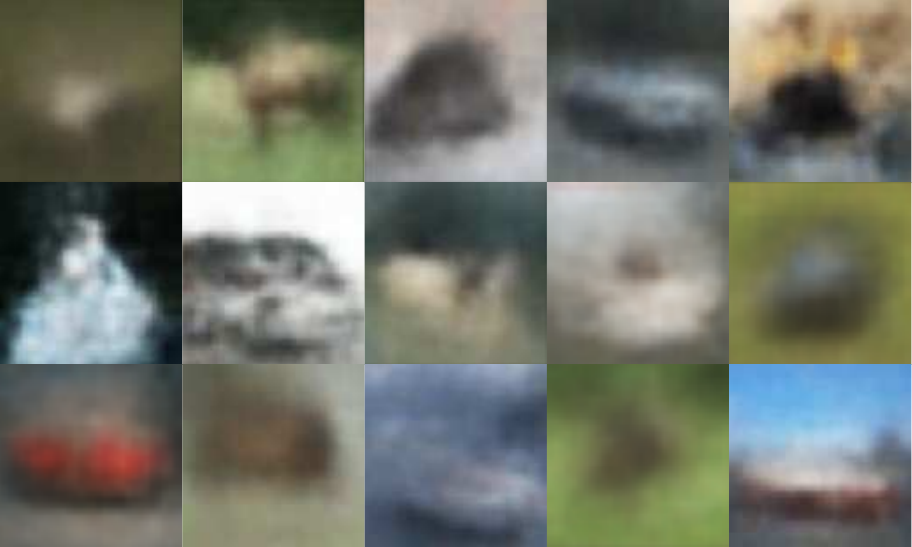}
                 & \includegraphics[clip,width=\linewidth]{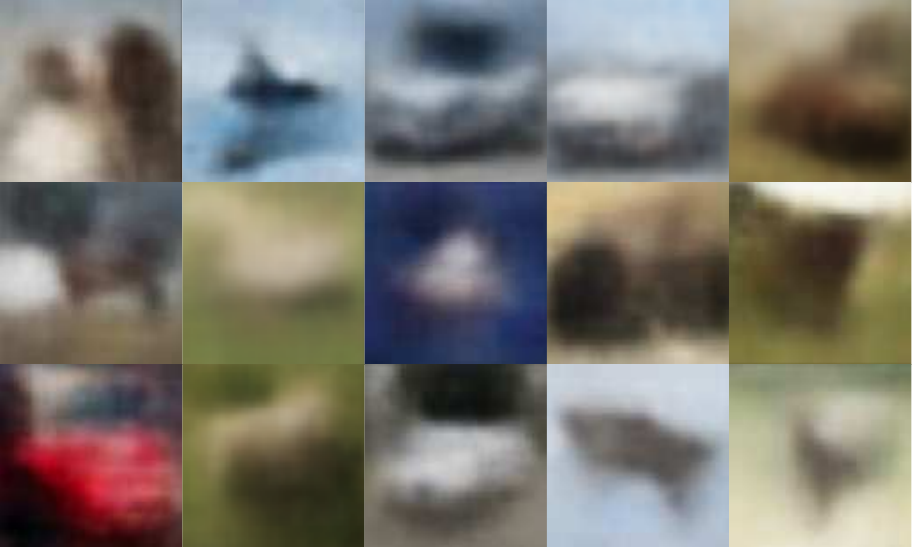}                 
                 & \includegraphics[clip,width=\linewidth]{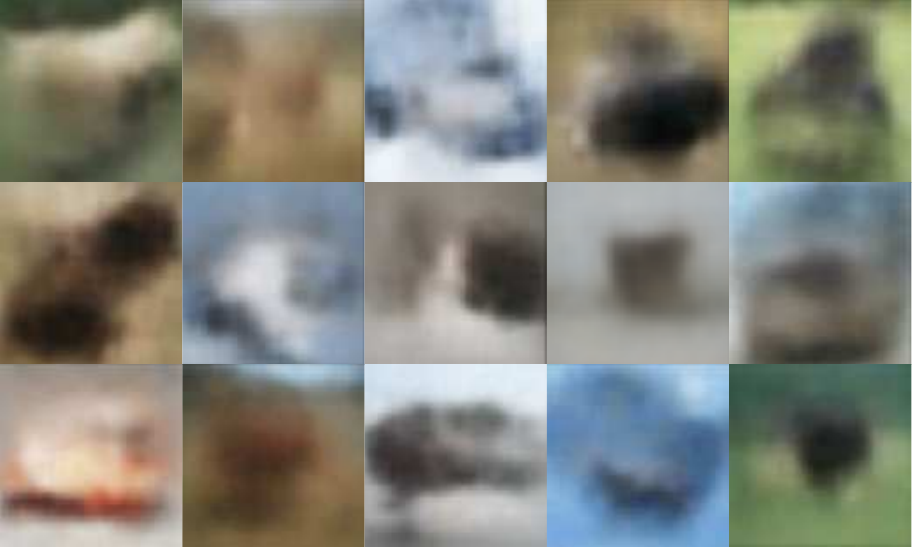}  \\

                 \midrule
                \raisebox{26pt}{$\beta$-vae}
                 & \includegraphics[clip,width=\linewidth]{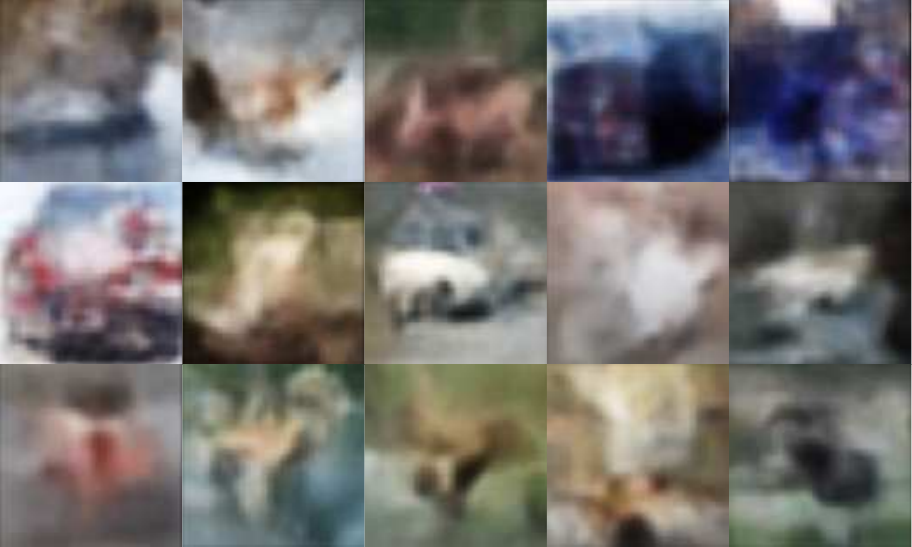}
                 & \includegraphics[clip,width=\linewidth]{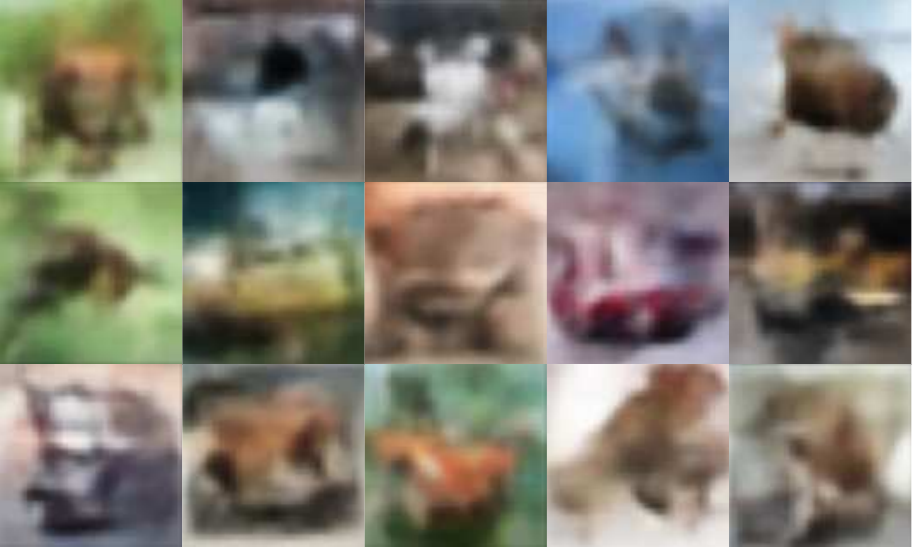}                 
                 & \includegraphics[clip,width=\linewidth]{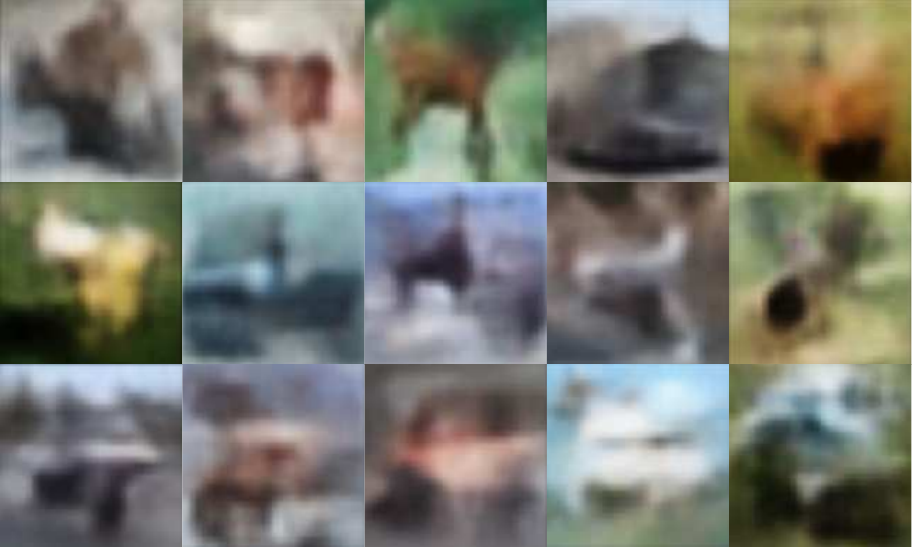}  \\

                 \midrule
                \raisebox{26pt}{hvae}
                 & \includegraphics[clip,width=\linewidth]{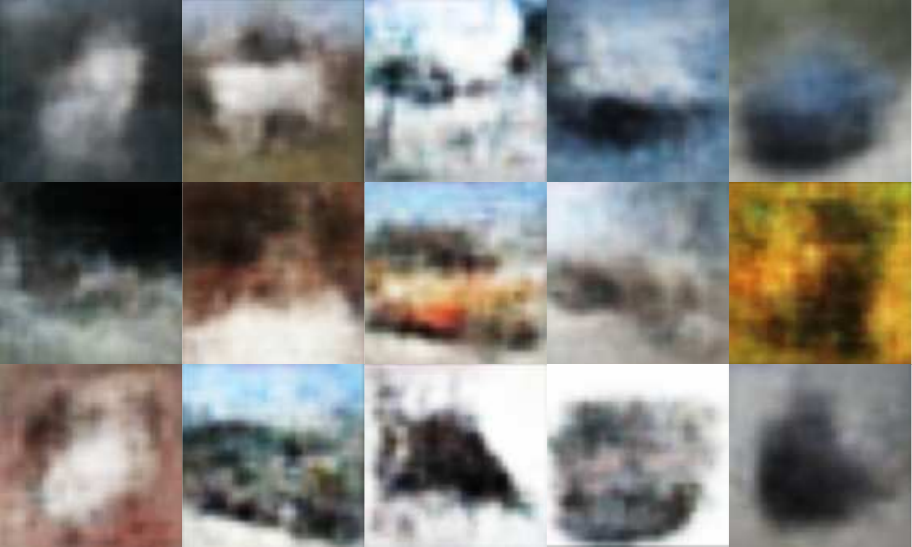}
                 & \includegraphics[clip,width=\linewidth]{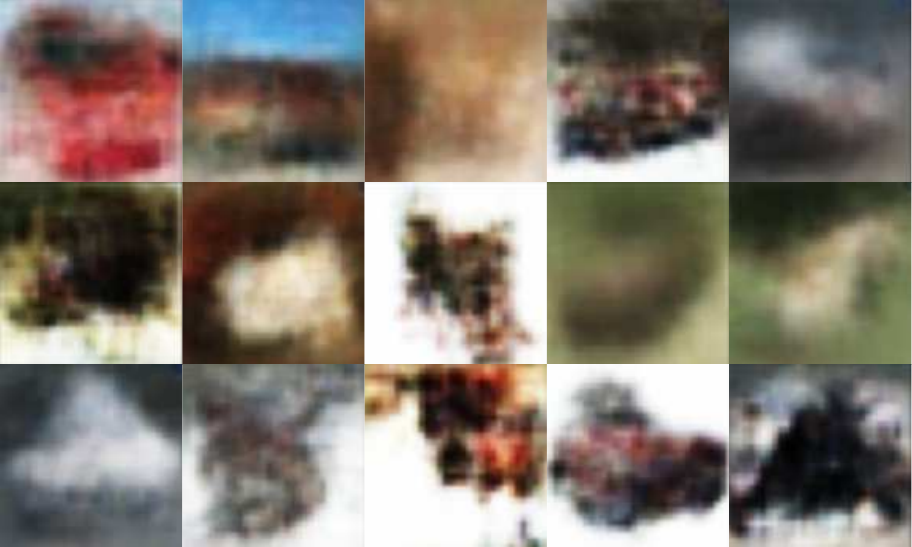}                 
                 & \includegraphics[clip,width=\linewidth]{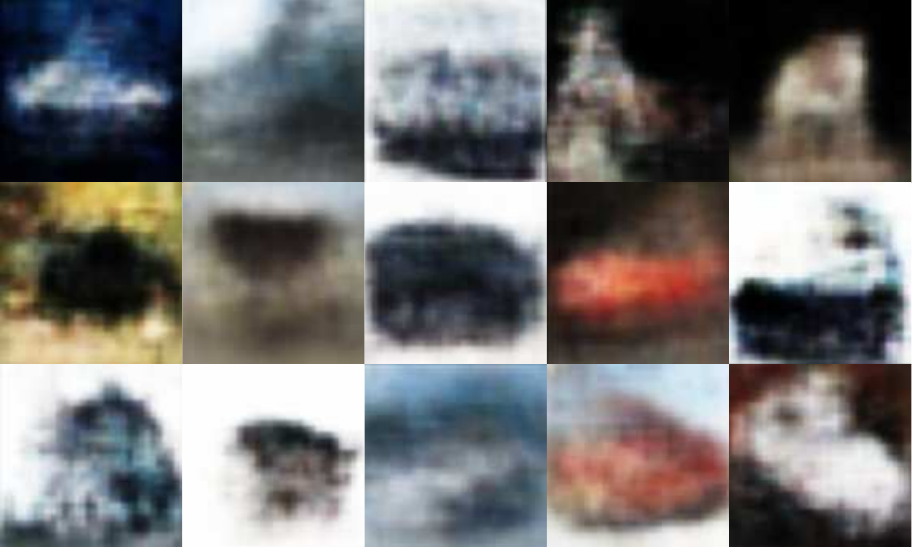}  \\
    
                \bottomrule
            \end{tabular}
        \end{sc}
    }
    \label{tab:cifar10_figures}
\end{table}

\newpage
\begin{table}[H]
    \caption{Uncurated samples from the models trained on the \celeba dataset.}
    \vspace{1ex}

    \centering

    \resizebox{\columnwidth}{!}{
        \setlength{\tabcolsep}{2pt}
        \begin{sc}
            \small

            \begin{tabular}{c  p{0.27\linewidth} p{0.27\linewidth} p{0.27\linewidth}}
                \toprule
                 & Vanilla & Gauss. Mollification & Blur. Mollification \\
                \midrule
                \midrule
                \raisebox{26pt}{real-nvp}
                 & \includegraphics[clip,width=\linewidth]{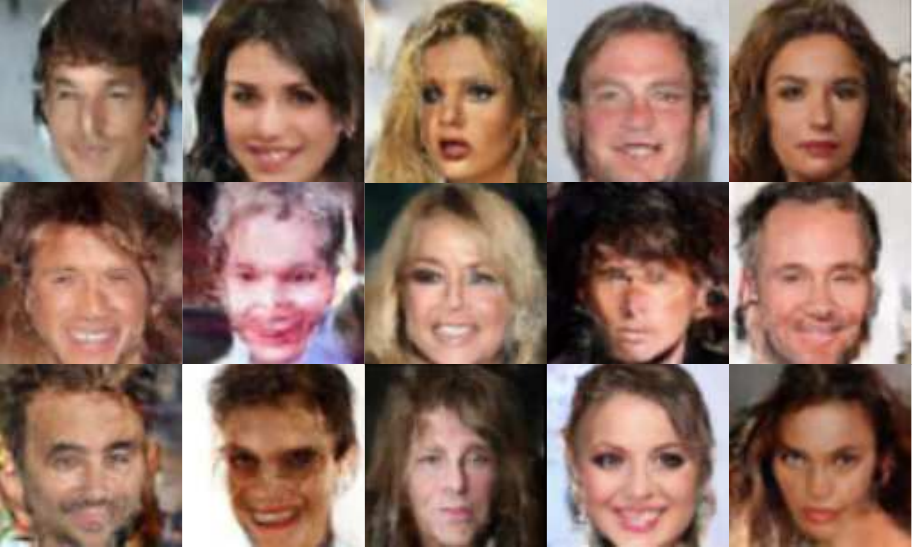}
                 & \includegraphics[clip,width=\linewidth]{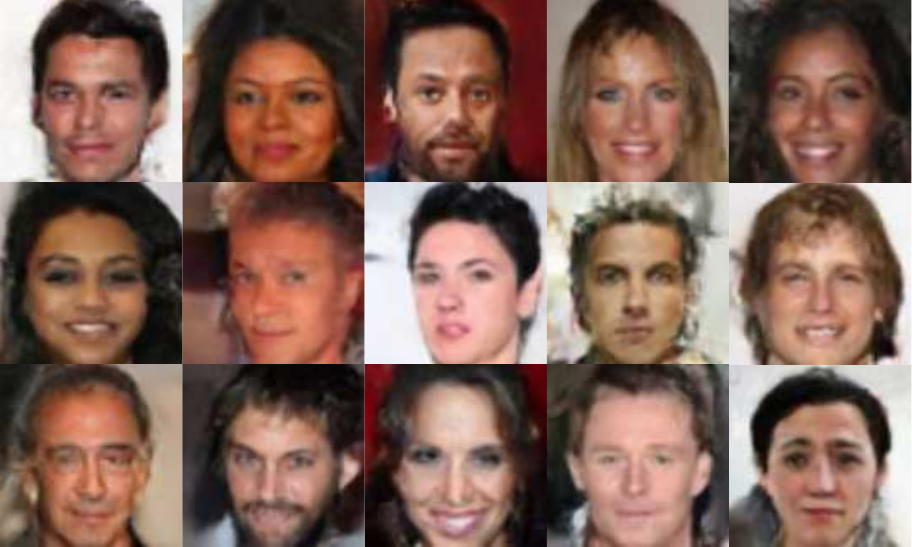}                 
                 & \includegraphics[clip,width=\linewidth]{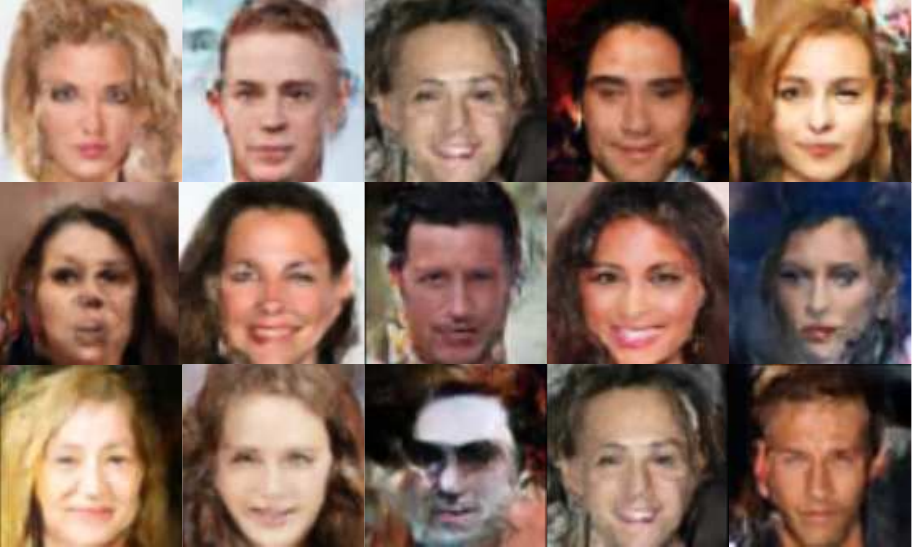}  \\

                \midrule
                \raisebox{26pt}{glow}
                 & \includegraphics[clip,width=\linewidth]{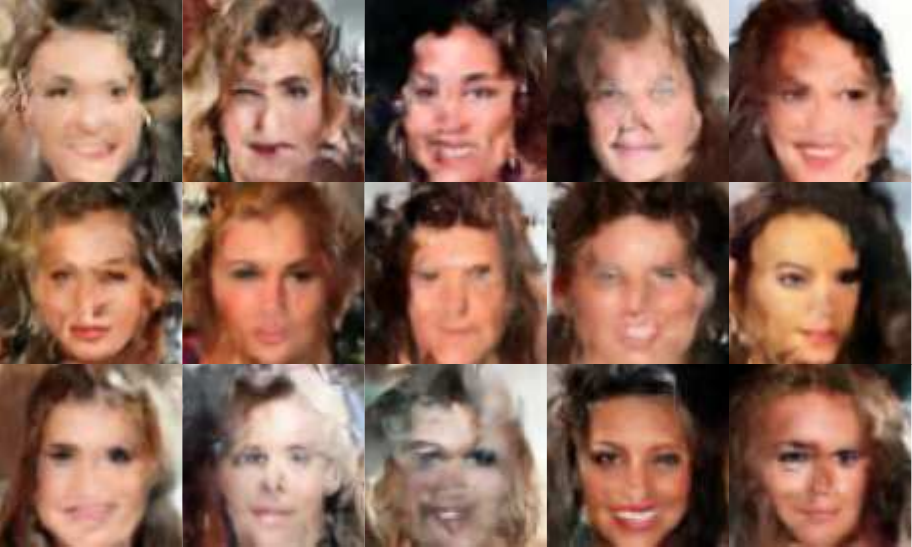}
                 & \includegraphics[clip,width=\linewidth]{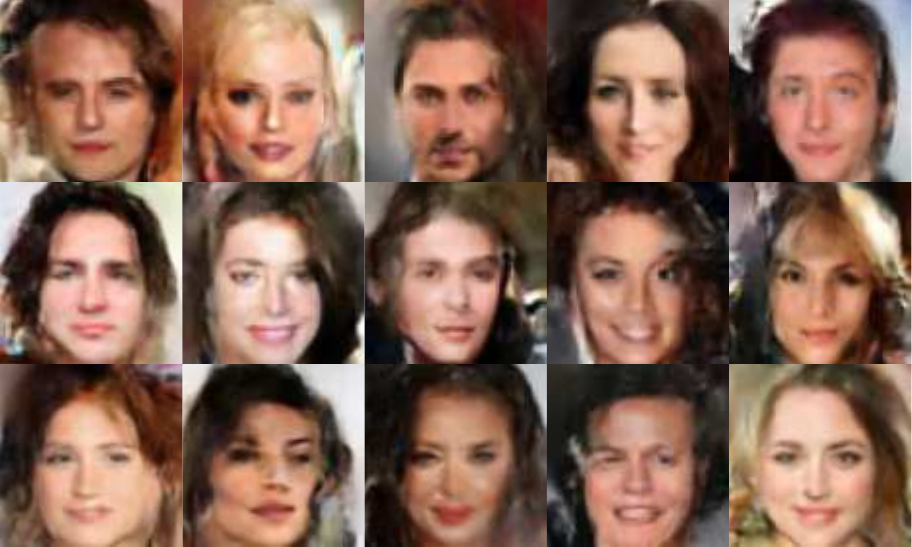}                 
                 & \includegraphics[clip,width=\linewidth]{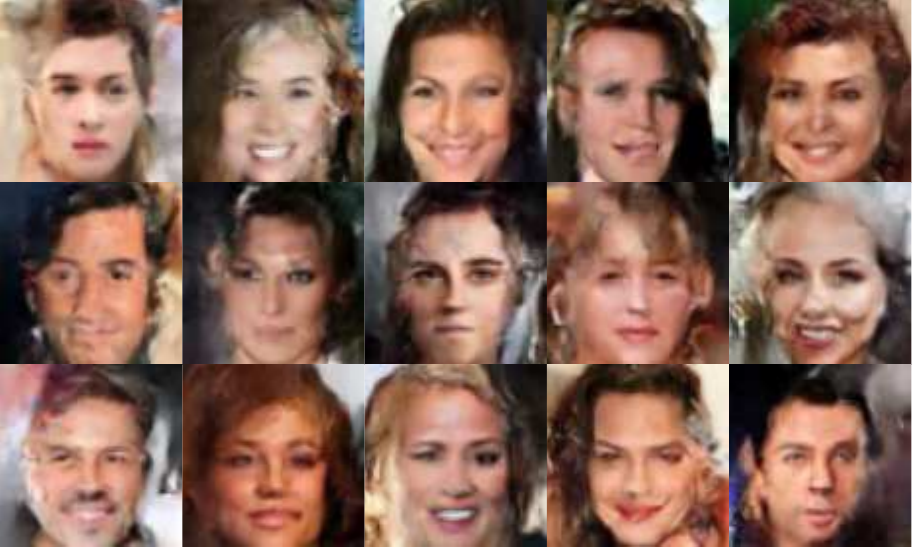}  \\

                 \midrule
                \raisebox{26pt}{vae}
                 & \includegraphics[clip,width=\linewidth]{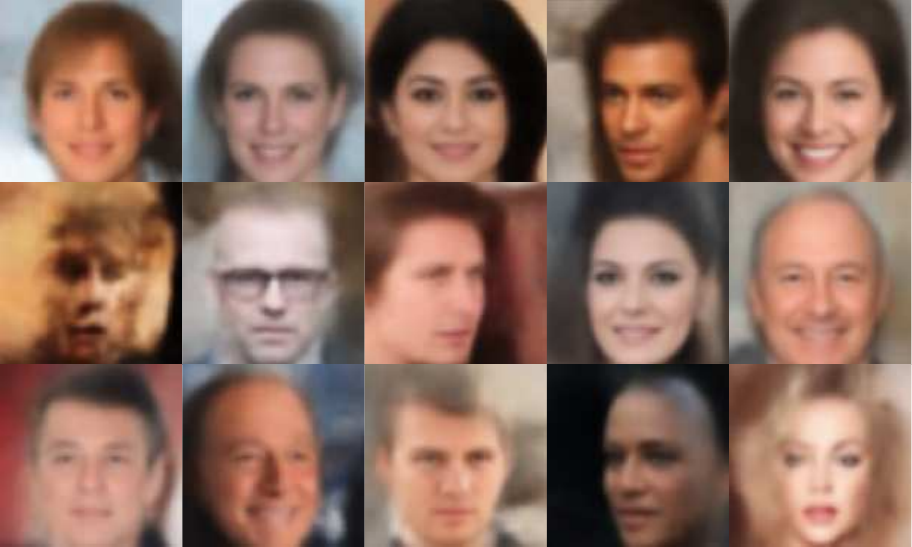}
                 & \includegraphics[clip,width=\linewidth]{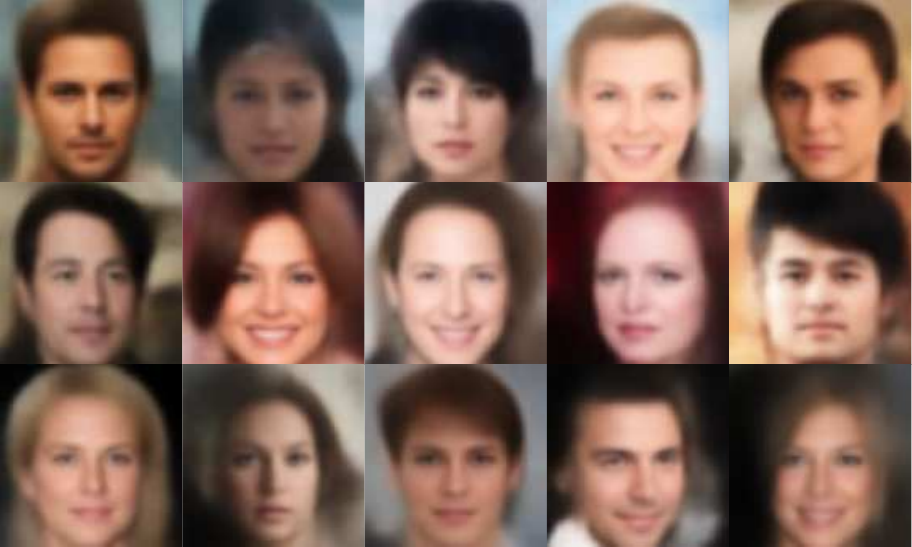}                 
                 & \includegraphics[clip,width=\linewidth]{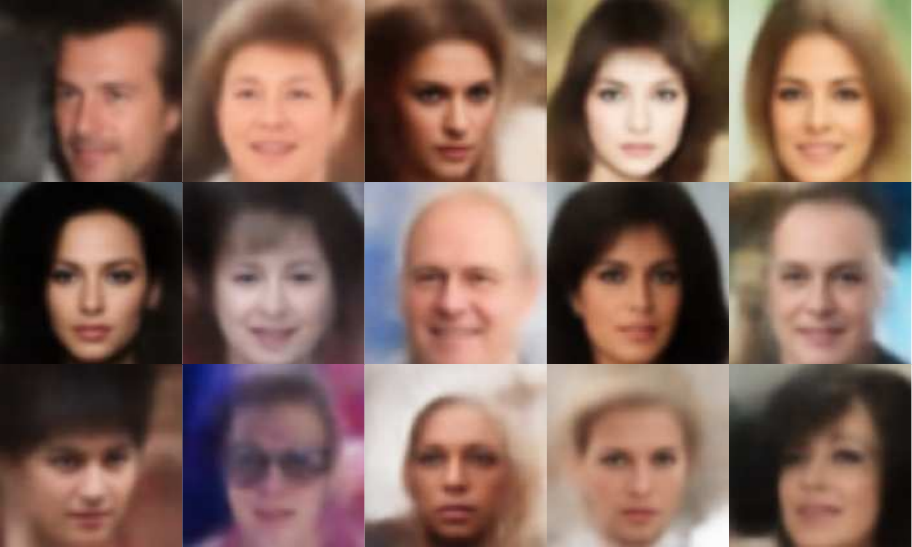}  \\

                 \midrule
                \raisebox{26pt}{vae-iaf}
                 & \includegraphics[clip,width=\linewidth]{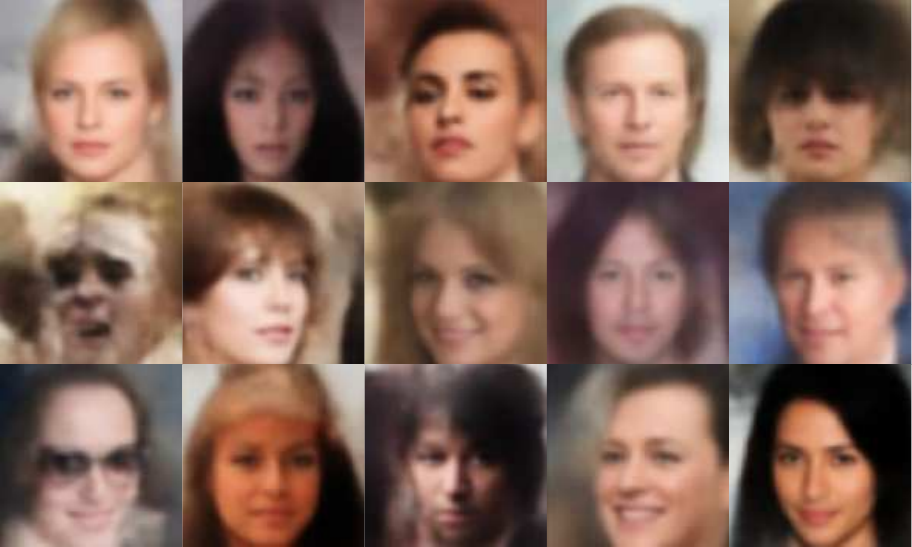}
                 & \includegraphics[clip,width=\linewidth]{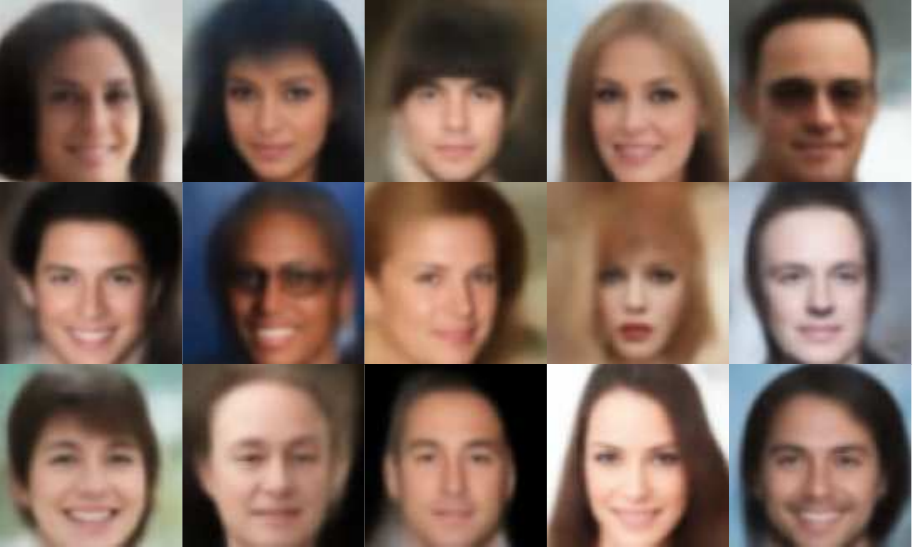}                 
                 & \includegraphics[clip,width=\linewidth]{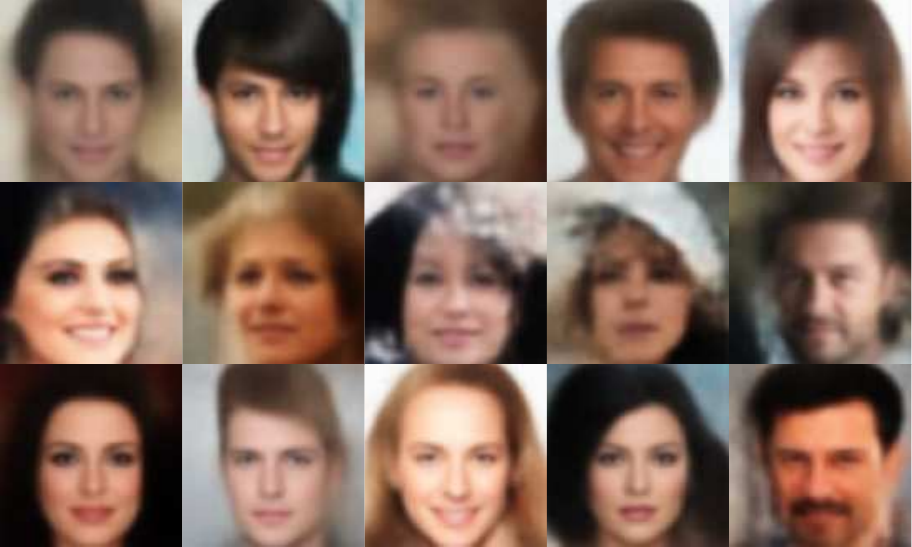}  \\

                 \midrule
                \raisebox{26pt}{iwae}
                 & \includegraphics[clip,width=\linewidth]{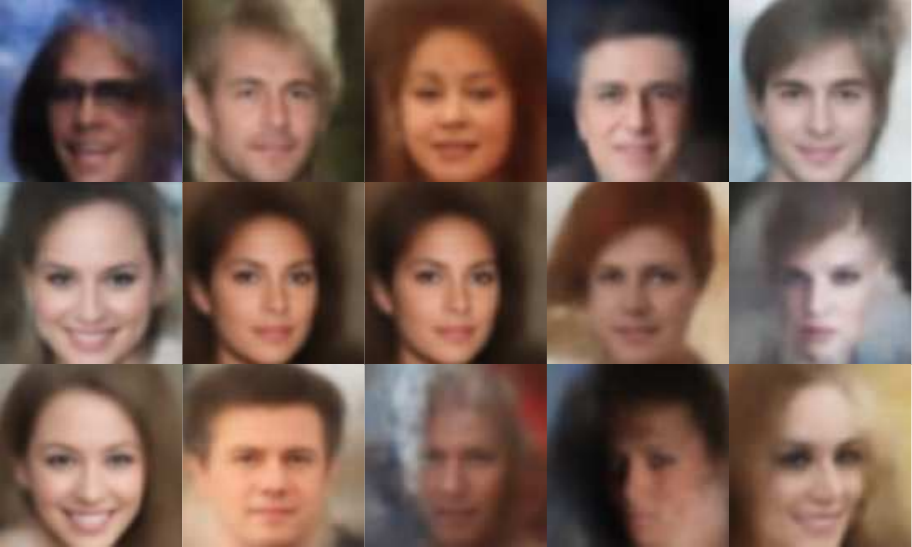}
                 & \includegraphics[clip,width=\linewidth]{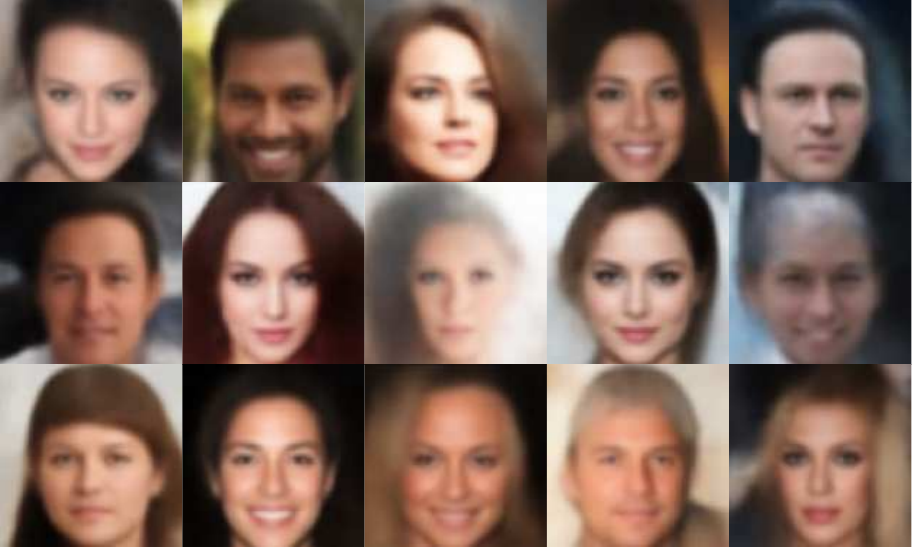}                 
                 & \includegraphics[clip,width=\linewidth]{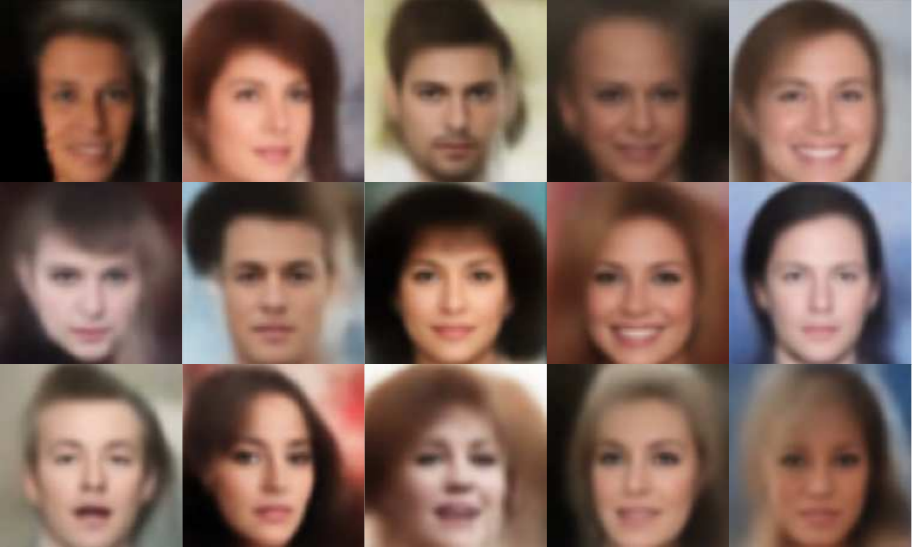}  \\

                 \midrule
                \raisebox{26pt}{$\beta$-vae}
                 & \includegraphics[clip,width=\linewidth]{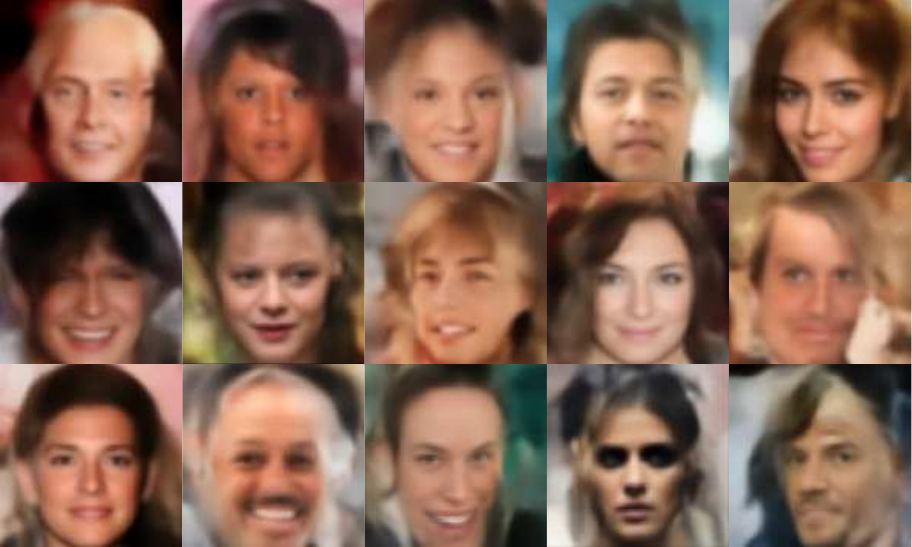}
                 & \includegraphics[clip,width=\linewidth]{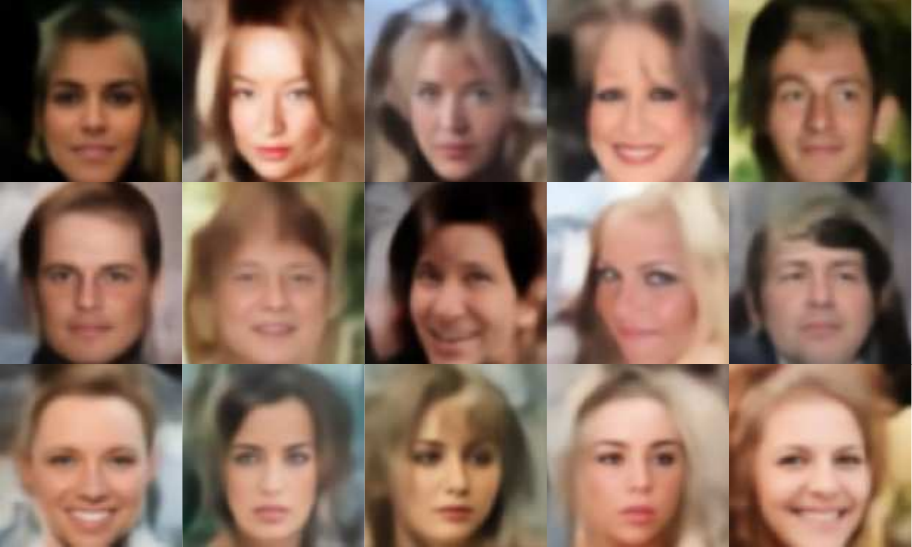}                 
                 & \includegraphics[clip,width=\linewidth]{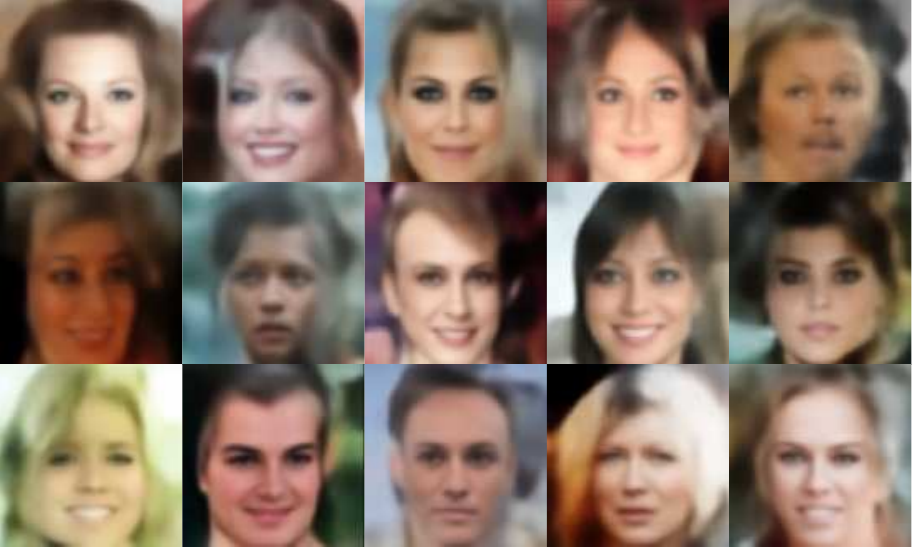}  \\

                 \midrule
                \raisebox{26pt}{hvae}
                 & \includegraphics[clip,width=\linewidth]{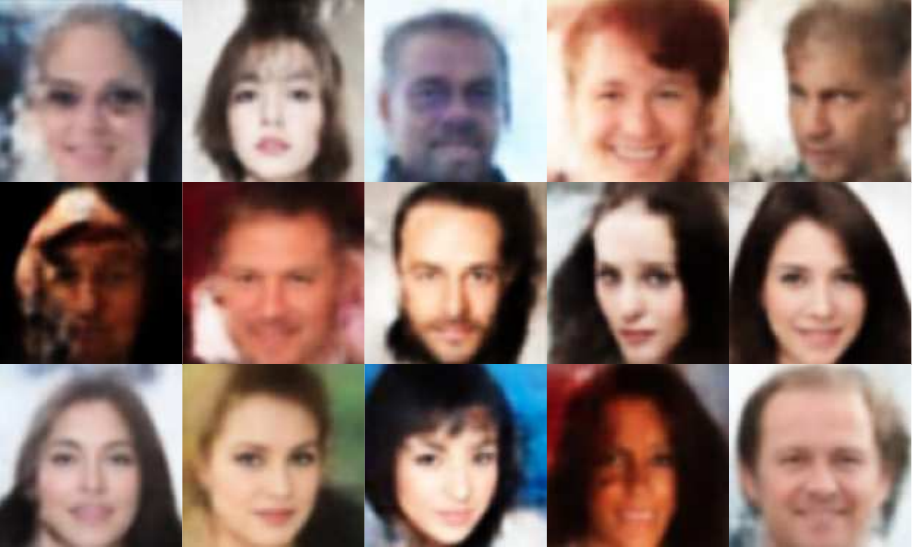}
                 & \includegraphics[clip,width=\linewidth]{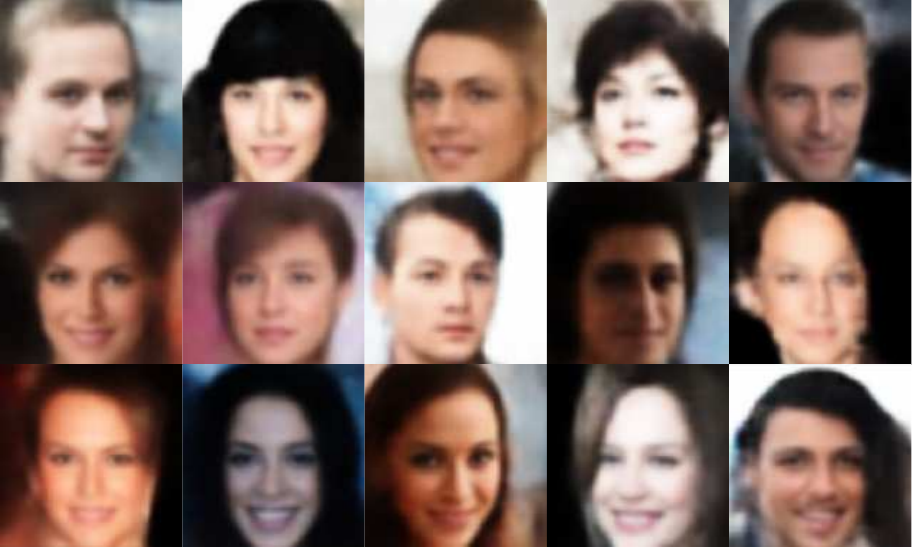}                 
                 & \includegraphics[clip,width=\linewidth]{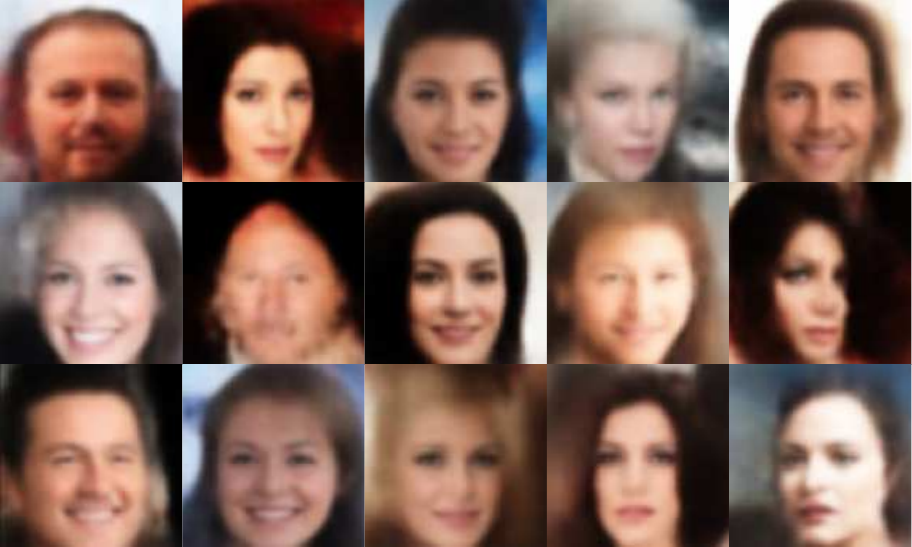}  \\
    
                \bottomrule
            \end{tabular}
        \end{sc}
    }
    \label{tab:celeba_figures}
\end{table}

\newpage

\begin{figure}[H]
    \centering
    \tikzexternaldisable
    \scriptsize
    \centering
    \setlength{\figurewidth}{5.2cm}
    \setlength{\figureheight}{4.7cm}
    \begin{subfigure}{0.3\textwidth}
        \begin{tikzpicture}

\definecolor{darkgray176}{RGB}{176,176,176}
\definecolor{firebrick}{RGB}{178,34,34}
\definecolor{seagreen}{RGB}{46,139,87}
\definecolor{steelblue}{RGB}{70,130,180}

\begin{axis}[
height=\figureheight,
major tick length=1ex, ymode=log,
tick align=outside,
tick pos=left,
title={VAE},
width=\figurewidth,
x grid style={darkgray176},
xlabel={Epoch},
xmin=-0.550000000000001, xmax=209.55,
xtick style={color=black},
y grid style={darkgray176},
ylabel={FID [log] (\(\displaystyle \leftarrow\))},
ymin=150, ymax=400,
ytick style={color=black}
]
\addplot [very thick, steelblue]
table {%
9 275.828360458979
19 258.475466804566
29 247.27933457069
39 243.317551677499
49 232.135458295369
59 222.347000737537
69 211.870721571321
79 205.543648883099
89 211.264828828225
99 205.192069721024
109 203.00336635333
119 201.409334983278
129 197.598290246665
139 194.017035167594
149 194.559061784145
159 191.942091427875
169 193.374292980827
179 190.477099855304
189 192.669358361209
200 191.986595275903
};
\addplot [very thick, firebrick]
table {%
9 284.995885067403
19 203.668304601768
29 185.739520570731
39 181.499296673885
49 183.763396123353
59 180.733472815963
69 173.470436966033
79 169.069473691144
89 169.708200350771
99 164.598705842794
109 163.932112087644
119 162.566692590737
129 160.904043429702
139 153.66002872769
149 159.640951948866
159 156.903666153181
169 154.67151302412
179 154.713318310392
189 153.000666174139
200 155.130112823793
};
\addplot [very thick, seagreen]
table {%
9 359.28524354459
19 383.134439210067
29 353.400773050816
39 376.956311211456
49 350.084101159933
59 298.390937790071
69 266.571382091969
79 239.090563455245
89 223.401217018332
99 209.097146174629
109 191.597088178865
119 187.980680791696
129 182.280596131415
139 176.236841885393
149 175.193717791509
159 171.662691253393
169 164.862513397538
179 165.17332998973
189 163.887022983427
200 162.695184159534
};
\end{axis}

\end{tikzpicture}
    \end{subfigure}
    \hfill
    \begin{subfigure}{0.3\textwidth}
        \begin{tikzpicture}

\definecolor{darkgray176}{RGB}{176,176,176}
\definecolor{firebrick}{RGB}{178,34,34}
\definecolor{seagreen}{RGB}{46,139,87}
\definecolor{steelblue}{RGB}{70,130,180}

\begin{axis}[
height=\figureheight,
major tick length=1ex, ymode=log,
tick align=outside,
tick pos=left,
title={\(\displaystyle \beta\)-VAE},
width=\figurewidth,
x grid style={darkgray176},
xlabel={Epoch},
xmin=-0.550000000000001, xmax=209.55,
xtick style={color=black},
y grid style={darkgray176},
ylabel={FID [log] (\(\displaystyle \leftarrow\))},
ymin=90, ymax=400,
ytick style={color=black}
]
\addplot [very thick, steelblue]
table {%
9 159.232738686969
19 133.119915748248
29 123.601114650319
39 115.60421355295
49 111.082401253058
59 111.936882133853
69 112.611061423112
79 111.326657218311
89 110.923642955505
99 111.260535606737
109 111.566795223487
119 111.473248450513
129 112.351132089474
139 108.948950353057
149 112.141846213607
159 110.650793570693
169 108.761463226362
179 110.253765314369
189 109.508868546068
200 112.427413952725
};
\addplot [very thick, firebrick]
table {%
9 409.17297852005
19 395.355426156324
29 380.796061070748
39 366.004288156073
49 328.125402585661
59 307.07277516818
69 150.738938167823
79 118.333106991288
89 103.832424607476
99 100.025636329481
109 99.0874678915784
119 98.577091353881
129 96.7807485914256
139 95.0764563409831
149 95.5869091985153
159 95.3102413864174
169 94.0301547342252
179 93.3976641318405
189 93.3625662113038
200 93.9042993846672
};
\addplot [very thick, seagreen]
table {%
9 356.537479482792
19 344.184819706719
29 334.013389702466
39 316.585755940079
49 279.879789859123
59 232.602666566201
69 200.488563914225
79 167.375518243453
89 149.650681048567
99 134.135605248096
109 119.022116782454
119 114.37833195258
129 108.26736896969
139 104.013235408148
149 100.994247394254
159 102.340073981804
169 99.2515360123799
179 102.581817701901
189 100.935143269171
200 101.301919318413
};
\end{axis}

\end{tikzpicture}
    \end{subfigure}
    \hfill
    \begin{subfigure}{0.3\textwidth}
        \begin{tikzpicture}

\definecolor{darkgray176}{RGB}{176,176,176}
\definecolor{firebrick}{RGB}{178,34,34}
\definecolor{seagreen}{RGB}{46,139,87}
\definecolor{steelblue}{RGB}{70,130,180}

\begin{axis}[
height=\figureheight,
major tick length=1ex, ymode=log,
tick align=outside,
tick pos=left,
title={IWAE},
width=\figurewidth,
x grid style={darkgray176},
xlabel={Epoch},
xmin=-0.550000000000001, xmax=209.55,
xtick style={color=black},
y grid style={darkgray176},
ylabel={FID [log] (\(\displaystyle \leftarrow\))},
ymin=140, ymax=400,
ytick style={color=black}
]
\addplot [very thick, steelblue]
table {%
9 273.918514198442
19 244.144917975096
29 238.594570354378
39 225.727022543827
49 211.489447776677
59 207.787795114677
69 204.379258097153
79 197.213408155923
89 192.396221011206
99 192.986181920704
109 189.231394385563
119 188.17586580772
129 185.499969541636
139 189.580938355063
149 188.881339754212
159 186.051638372946
169 184.033983752531
179 182.668680199297
189 180.200160890868
200 183.049035660565
};
\addplot [very thick, firebrick]
table {%
9 322.917175078091
19 286.990099256275
29 250.999409550674
39 227.602720504852
49 220.601692827405
59 201.871725045675
69 189.966809478052
79 177.718621084305
89 173.215156633161
99 168.449169666205
109 162.534380166565
119 161.160073913933
129 159.510879236315
139 153.753291810624
149 155.41758836198
159 150.7349875802
169 149.172409249532
179 148.916938479745
189 146.274088704605
200 146.708287192323
};
\addplot [very thick, seagreen]
table {%
9 377.40394521019
19 396.798820870272
29 386.940186026397
39 367.576570053305
49 345.769347499451
59 314.877050537448
69 277.259441042219
79 246.5309872368
89 227.048235868159
99 212.5112399149
109 198.042448557296
119 195.358945399139
129 186.146453327378
139 178.546380127597
149 177.665032342543
159 172.097132890439
169 166.43578712623
179 165.428278395442
189 162.90537920456
200 163.797319725828
};
\end{axis}

\end{tikzpicture}
    \end{subfigure}

    \begin{subfigure}{0.3\textwidth}
        \begin{tikzpicture}

\definecolor{darkgray176}{RGB}{176,176,176}
\definecolor{firebrick}{RGB}{178,34,34}
\definecolor{seagreen}{RGB}{46,139,87}
\definecolor{steelblue}{RGB}{70,130,180}

\begin{axis}[
height=\figureheight,
major tick length=1ex, ymode=log,
tick align=outside,
tick pos=left,
title={VAE-IAF},
width=\figurewidth,
x grid style={darkgray176},
xlabel={Epoch},
xmin=-0.550000000000001, xmax=209.55,
xtick style={color=black},
y grid style={darkgray176},
ylabel={FID [log] (\(\displaystyle \leftarrow\))},
ymin=150, ymax=400,
ytick style={color=black}
]
\addplot [very thick, steelblue]
table {%
9 263.58930946813
19 261.004724694414
29 240.172889574185
39 243.996962866979
49 229.374473450733
59 230.783509938168
69 213.522824493911
79 216.192530210501
89 213.710031165255
99 210.946179374731
109 205.939523649922
119 201.011413698497
129 198.721717947189
139 199.739795462252
149 200.882283785928
159 197.501567765188
169 197.513953841826
179 197.691810245957
189 195.65644189365
200 193.058221445748
};
\addplot [very thick, firebrick]
table {%
9 343.898303098139
19 288.141768445986
29 243.139559053628
39 229.421658517983
49 216.687201605929
59 202.172513143573
69 196.435209623394
79 190.437595998513
89 182.129609690526
99 178.939134176881
109 175.098646858922
119 171.695552862326
129 165.725438780811
139 164.727540483803
149 164.073101218026
159 161.32807813927
169 157.682758824269
179 156.817297984346
189 158.078195565075
200 156.399442921092
};
\addplot [very thick, seagreen]
table {%
9 377.28114288876
19 384.104907135015
29 357.012380812062
39 348.649019656504
49 316.339855744542
59 269.138314143718
69 241.93282632222
79 221.099305854718
89 212.279682693372
99 199.633263530653
109 189.132759986286
119 185.834855859892
129 179.344652889708
139 174.233865785947
149 173.113018487308
159 170.266300285226
169 168.456989164223
179 169.549348160132
189 165.923646945972
200 162.271606750238
};
\end{axis}

\end{tikzpicture}
    \end{subfigure}
    \hfill
    \begin{subfigure}{0.3\textwidth}
        \begin{tikzpicture}

\definecolor{darkgray176}{RGB}{176,176,176}
\definecolor{firebrick}{RGB}{178,34,34}
\definecolor{seagreen}{RGB}{46,139,87}
\definecolor{steelblue}{RGB}{70,130,180}

\begin{axis}[
height=\figureheight,
major tick length=1ex, ymode=log,
tick align=outside,
tick pos=left,
title={HVAE},
width=\figurewidth,
x grid style={darkgray176},
xlabel={Epoch},
xmin=-0.550000000000001, xmax=209.55,
xtick style={color=black},
y grid style={darkgray176},
ylabel={FID [log] (\(\displaystyle \leftarrow\))},
ymin=135, ymax=400,
ytick style={color=black}
]
\addplot [very thick, steelblue]
table {%
9 291.663585574702
19 261.473904054148
29 261.885321717575
39 239.52284285876
49 230.656494980545
59 213.321838639689
69 215.277393308126
79 199.420543359482
89 196.965028222947
99 200.884051507038
109 196.728660955017
119 187.139937241524
129 183.141156274645
139 181.07625886556
149 183.068269501783
159 176.838247703296
169 182.783420999622
179 170.818330545268
189 169.311947797988
200 173.472020233385
};
\addplot [very thick, firebrick]
table {%
9 312.572130242788
19 306.175522873166
29 279.510629742975
39 263.495172026817
49 245.567833545639
59 221.866730398843
69 206.484837796381
79 192.756001418587
89 184.77634060702
99 178.019107369936
109 166.543520548116
119 167.922002038732
129 159.792291339619
139 153.257051244254
149 153.511047823731
159 146.302970616601
169 141.677333723136
179 143.915642812405
189 141.651953949318
200 137.844908924975
};
\addplot [very thick, seagreen]
table {%
9 358.509166665584
19 366.955301882325
29 337.671142292162
39 342.503466360803
49 318.15055177141
59 280.273725461318
69 259.007864895027
79 233.219793141438
89 210.838210728355
99 201.222733119176
109 188.453029709193
119 173.982641050281
129 172.718880747559
139 167.239386156797
149 164.122630750107
159 157.51882305846
169 156.541560763982
179 153.580608897205
189 149.107803682666
200 147.151040441363
};
\end{axis}

\end{tikzpicture}
    \end{subfigure}
    \hfill
    \begin{subfigure}{0.3\textwidth}
        \begin{tikzpicture}

\definecolor{darkgray176}{RGB}{176,176,176}
\definecolor{firebrick}{RGB}{178,34,34}
\definecolor{seagreen}{RGB}{46,139,87}
\definecolor{steelblue}{RGB}{70,130,180}

\begin{axis}[
height=\figureheight,
major tick length=1ex,
tick align=outside,
tick pos=left,
width=\figurewidth,
x grid style={darkgray176},
xlabel={Epoch},
xmin=-2.95, xmax=83.95,
xtick style={color=black},
y grid style={darkgray176},
ylabel={FID [log] (\(\displaystyle \leftarrow\))},
title={GLOW},
ymode=log,
ymin=60, ymax=440,
ytick style={color=black}
]
\addplot [very thick, steelblue]
table {%
1 219.87905533115
2 173.191181029243
4 132.807374023949
5 122.562572364224
7 106.853497449081
9 102.342958953606
10 90.5040287142261
12 91.7662795858768
13 87.8931117376714
15 89.9819737912088
17 81.2961696439393
18 79.8161413690451
20 80.3415839962055
21 81.2312359009489
23 79.852354177318
25 80.1597453835761
26 78.7095287263228
28 76.8321190404832
30 78.9191622234901
31 78.2146108053648
33 80.0399382814894
34 80.5643798520238
36 78.6578736525902
38 78.9095321516281
39 76.8228420780823
41 77.7902512388748
42 78.5054324951778
44 75.5641108620846
46 76.3896550518472
47 76.1175611462584
49 79.6537578746575
50 76.4976338942581
52 77.9594513673208
54 75.5071758217979
55 76.2651715831182
57 76.7448804506168
59 75.8643348810895
60 75.1648242279502
62 77.3130080543195
63 75.9836639438986
65 75.698175954769
67 76.8550461015116
68 77.0208939817639
70 75.5197660572222
71 77.0376830954293
73 73.8980899895075
75 73.7943582061472
76 75.9738976074497
78 76.4807633695735
80 74.6266966989407
};
\addplot [very thick, firebrick]
table {%
1 434.259901707134
2 430.817218383184
4 420.210304277221
5 404.416296951276
7 395.01995870988
9 383.083867296336
10 368.712904477546
12 358.622953604526
13 348.183697283412
15 333.016386483374
17 328.098336521876
18 318.191993476283
20 307.647145662772
21 296.042305131146
23 284.969471971731
25 282.502110738499
26 276.041382367121
28 271.194811842369
30 261.071130497678
31 252.882721725255
33 249.074045275904
34 238.09344890853
36 232.048267991833
38 218.144833914201
39 213.114928036612
41 202.213959405462
42 187.270615480434
44 175.044414854995
46 148.535906321917
47 89.8168670518874
49 76.7439963468889
50 74.1820498508703
52 69.6880934836911
54 69.9143271479604
55 68.0051366495035
57 67.1408933526313
59 66.3872527106699
60 68.3246358922891
62 67.952158344012
63 66.7338470531034
65 66.3287869939228
67 68.5069074628458
68 66.6535026567948
70 67.0255801531943
71 65.6370019018968
73 64.8816345636447
75 66.0719062518673
76 64.830590898642
78 66.4892999039883
80 64.8788928186547
};
\addplot [very thick, seagreen]
table {%
1 306.806753698433
2 367.854737605121
4 354.609023435558
5 353.916827707974
7 399.726268439631
9 429.128394934453
10 412.110150386918
12 403.436322395884
13 395.361689967761
15 373.058248915857
17 348.00470100772
18 304.45703821484
20 273.595084231463
21 249.771735853673
23 229.551222448051
25 226.889186110185
26 214.485662528359
28 203.929066563399
30 194.391242739334
31 189.508004103682
33 185.930704040347
34 190.602166984679
36 176.566396554872
38 156.82674875017
39 149.524483606601
41 157.946103661143
42 126.31331121423
44 110.406804467264
46 104.532674661716
47 93.1481007848361
49 97.8761246065096
50 88.3844309070877
52 80.2218248330981
54 76.8936127838215
55 78.4675253367593
57 77.8098616388431
59 73.7843618506092
60 71.0150231408294
62 72.0647587974309
63 69.8286146112885
65 72.2446970756613
67 71.7477855029989
68 69.344385847102
70 69.1409830725254
71 68.0282119467219
73 69.7985872284308
75 72.2093027720322
76 68.3974138040853
78 67.1394089982266
80 67.2945465828857
};
\end{axis}

\end{tikzpicture}
    \end{subfigure}

    \tikzexternalenable

    \caption{The progression of FID scores during training on the \cifar dataset. \label{fig:cifar_fid_app}}
\end{figure}

\begin{figure}[H]
    \centering
    \tikzexternaldisable
    \scriptsize
    \centering
    \setlength{\figurewidth}{5.2cm}
    \setlength{\figureheight}{4.7cm}
    \begin{subfigure}{0.3\textwidth}
        \begin{tikzpicture}

\definecolor{darkgray176}{RGB}{176,176,176}
\definecolor{firebrick}{RGB}{178,34,34}
\definecolor{seagreen}{RGB}{46,139,87}
\definecolor{steelblue}{RGB}{70,130,180}

\begin{axis}[
height=\figureheight,
major tick length=1ex, ymode=log,
tick align=outside,
tick pos=left,
title={VAE},
width=\figurewidth,
x grid style={darkgray176},
xlabel={Epoch},
xmin=4.45, xmax=104.55,
xtick style={color=black},
y grid style={darkgray176},
ylabel={FID [log] (\(\displaystyle \leftarrow\))},
ymin=70, ymax=400,
ytick style={color=black}
]
\addplot [very thick, steelblue]
table {%
9 106.039637601177
19 97.5644019435999
29 91.1023771020631
39 88.6507213854308
49 86.1080635652987
59 84.5825436778091
69 81.8782913679911
79 81.6889544785797
89 80.9740660481404
100 80.1907309288866
};
\addplot [very thick, firebrick]
table {%
9 222.804409356609
19 95.7236463393244
29 88.1286691010594
39 84.6450434675845
49 82.9230527662934
59 79.896201262105
69 78.1200751187392
79 76.5608443161193
89 74.1939811864941
100 72.9719703351568
};
\addplot [very thick, seagreen]
table {%
9 426.674307191818
19 338.172702546759
29 178.376478517755
39 115.69132427323
49 91.3346630286292
59 83.8798857466387
69 82.3826637352498
79 80.0398614951486
89 79.0495064351985
100 77.2903257528053
};
\end{axis}

\end{tikzpicture}
    \end{subfigure}
    \hfill
    \begin{subfigure}{0.3\textwidth}
        \begin{tikzpicture}

\definecolor{darkgray176}{RGB}{176,176,176}
\definecolor{firebrick}{RGB}{178,34,34}
\definecolor{seagreen}{RGB}{46,139,87}
\definecolor{steelblue}{RGB}{70,130,180}

\begin{axis}[
height=\figureheight,
major tick length=1ex, ymode=log,
tick align=outside,
tick pos=left,
title={\(\displaystyle \beta\)-VAE},
width=\figurewidth,
x grid style={darkgray176},
xlabel={Epoch},
xmin=4.45, xmax=104.55,
xtick style={color=black},
y grid style={darkgray176},
ylabel={FID [log] (\(\displaystyle \leftarrow\))},
ymin=60, ymax=400,
ytick style={color=black}
]
\addplot [very thick, steelblue]
table {%
9 79.1308635557261
19 75.1487168911413
29 72.9953981188444
39 71.5062640723687
49 70.8229637066536
59 69.5441705224847
69 69.4431370069188
79 69.0551142756996
89 68.3244852655226
100 67.7890878336757
};
\addplot [very thick, firebrick]
table {%
9 339.786295400022
19 371.841467718397
29 359.566831644305
39 80.6445357970172
49 71.1599324371961
59 68.3654456702052
69 66.7136329303064
79 66.0351319230093
89 65.1187528114363
100 64.5915745147402
};
\addplot [very thick, seagreen]
table {%
9 427.212964017876
19 311.54979192356
29 176.641531930583
39 99.722850177711
49 76.6642250731276
59 72.1726745072463
69 69.848812046813
79 68.988545785876
89 67.9684102453377
100 67.0891280590503
};
\end{axis}

\end{tikzpicture}
    \end{subfigure}
    \hfill
    \begin{subfigure}{0.3\textwidth}
        \begin{tikzpicture}

\definecolor{darkgray176}{RGB}{176,176,176}
\definecolor{firebrick}{RGB}{178,34,34}
\definecolor{seagreen}{RGB}{46,139,87}
\definecolor{steelblue}{RGB}{70,130,180}

\begin{axis}[
height=\figureheight,
major tick length=1ex, ymode=log,
tick align=outside,
tick pos=left,
title={IWAE},
width=\figurewidth,
x grid style={darkgray176},
xlabel={Epoch},
xmin=4.45, xmax=104.55,
xtick style={color=black},
y grid style={darkgray176},
ylabel={FID [log] (\(\displaystyle \leftarrow\))},
ymin=70, ymax=400,
ytick style={color=black}
]
\addplot [very thick, steelblue]
table {%
9 101.086265471211
19 94.6689824541269
29 87.7123105063191
39 86.0647383846374
49 82.7556729385281
59 81.711751735367
69 79.8438802087849
79 78.9344867462043
89 78.4064983761494
100 78.2521471117568
};
\addplot [very thick, firebrick]
table {%
9 124.077663481968
19 93.1990452880603
29 84.9167575326151
39 82.2527658388308
49 79.5627886859708
59 76.6004534297983
69 74.4818006489963
79 73.8177995052177
89 71.9355221959862
100 71.3871014353954
};
\addplot [very thick, seagreen]
table {%
9 438.778530561572
19 317.014023464368
29 177.330738462201
39 116.008724107459
49 88.8144152996461
59 82.2315642418013
69 81.0602651513061
79 79.3160745243933
89 77.5680925211055
100 76.4557590661307
};
\end{axis}

\end{tikzpicture}
    \end{subfigure}

    \begin{subfigure}{0.3\textwidth}
        \begin{tikzpicture}

\definecolor{darkgray176}{RGB}{176,176,176}
\definecolor{firebrick}{RGB}{178,34,34}
\definecolor{seagreen}{RGB}{46,139,87}
\definecolor{steelblue}{RGB}{70,130,180}

\begin{axis}[
height=\figureheight,
major tick length=1ex, ymode=log,
tick align=outside,
tick pos=left,
title={VAE-IAF},
width=\figurewidth,
x grid style={darkgray176},
xlabel={Epoch},
xmin=4.5, xmax=103.5,
xtick style={color=black},
y grid style={darkgray176},
ylabel={FID [log] (\(\displaystyle \leftarrow\))},
ymin=70, ymax=400,
ytick style={color=black}
]
\addplot [very thick, steelblue]
table {%
9 105.603124499453
19 97.8322736761228
29 91.6533095268834
39 89.5869772079062
49 86.2569474991851
59 84.7855409306132
69 83.1251504149903
79 81.7772378769743
89 80.8007255735402
99 80.8007255735402
};
\addplot [very thick, firebrick]
table {%
9 139.626711015902
19 101.597661448319
29 88.0619423765418
39 84.6648504467054
49 82.0100220376164
59 79.6686185770864
69 77.5130884901148
79 75.8457170143484
89 74.6954093906763
99 74.6954093906763
};
\addplot [very thick, seagreen]
table {%
9 437.185851853103
19 329.013791208025
29 181.428681565636
39 116.41746530874
49 92.7097059691246
59 84.4441096520735
69 82.0130152321561
79 79.3656519321712
89 78.0840512542929
99 78.0840512542929
};
\end{axis}

\end{tikzpicture}
    \end{subfigure}
    \hfill
    \begin{subfigure}{0.3\textwidth}
        \begin{tikzpicture}

\definecolor{darkgray176}{RGB}{176,176,176}
\definecolor{firebrick}{RGB}{178,34,34}
\definecolor{seagreen}{RGB}{46,139,87}
\definecolor{steelblue}{RGB}{70,130,180}

\begin{axis}[
height=\figureheight,
major tick length=1ex, ymode=log,
tick align=outside,
tick pos=left,
title={HVAE},
width=\figurewidth,
x grid style={darkgray176},
xlabel={Epoch},
xmin=4.45, xmax=104.55,
xtick style={color=black},
y grid style={darkgray176},
ylabel={FID [log] (\(\displaystyle \leftarrow\))},
ymin=70, ymax=400,
ytick style={color=black}
]
\addplot [very thick, steelblue]
table {%
9 103.635322414989
19 94.5092209639012
29 81.3576870641789
39 79.5129070516412
49 77.8610383384721
59 76.3611488584671
69 75.3519159115727
79 75.8252338571191
89 74.61936801736
100 74.1028906552917
};
\addplot [very thick, firebrick]
table {%
9 144.463778966184
19 97.0418195358193
29 88.1233158104204
39 81.9544394844948
49 78.9024719677694
59 75.0389883451949
69 74.3366791821662
79 72.5302892301422
89 72.5760086220234
100 72.4579616255297
};
\addplot [very thick, seagreen]
table {%
9 413.697197080173
19 349.476874911094
29 197.526511096574
39 121.490549684653
49 91.6042547385084
59 81.9310706217364
69 81.1991194940432
79 79.8053169683424
89 78.9396946501142
100 77.5460346621743
};
\end{axis}

\end{tikzpicture}
    \end{subfigure}
    \hfill
    \begin{subfigure}{0.3\textwidth}
        \begin{tikzpicture}

\definecolor{darkgray176}{RGB}{176,176,176}
\definecolor{firebrick}{RGB}{178,34,34}
\definecolor{seagreen}{RGB}{46,139,87}
\definecolor{steelblue}{RGB}{70,130,180}

\begin{axis}[
height=\figureheight,
major tick length=1ex, ymode=log,
tick align=outside,
tick pos=left,
title={GLOW},
width=\figurewidth,
x grid style={darkgray176},
xlabel={Epoch},
xmin=2.2, xmax=41.8,
xtick style={color=black},
y grid style={darkgray176},
ylabel={FID [log] (\(\displaystyle \leftarrow\))},
ymin=60, ymax=450,
ytick style={color=black}
]
\addplot [very thick, steelblue]
table {%
4 186.82523
6.4 125.960643
8.8 118.895994
11.2 121.487263
13.6 120.226265
16 102.877597
18.4 101.430122
20.8 104.081023
23.2 95.495871
25.6 95.633103
28 92.081574
30.4 99.515797
32.8 92.293891
35.2 95.016426
37.6 96.660847
40 97.597946
};
\addplot [very thick, firebrick]
table {%
4 447.176529
6.4 426.970754
8.8 392.383894
11.2 346.212662
13.6 331.919092
16 298.712465
18.4 226.87197
20.8 105.154286
23.2 103.633942
25.6 100.649619
28 83.224751
30.4 76.268658
32.8 72.477195
35.2 73.25018
37.6 70.717969
40 67.44256
};
\addplot [very thick, seagreen]
table {%
4 374.252514
6.4 575.987261
8.8 332.516392
11.2 364.75897
13.6 352.471131
16 348.069789
18.4 289.982549
20.8 169.385317
23.2 131.249644
25.6 109.50563
28 107.939149
30.4 102.058298
32.8 99.967368
35.2 100.786126
37.6 97.651416
40 94.10782
};
\end{axis}

\end{tikzpicture}
    \end{subfigure}

    \tikzexternalenable

    \caption{The progression of FID scores during training on the \celeba dataset. \label{fig:celeba_fid_app}}
\end{figure}

\end{document}